%% file: camera_ready.tex
\pdfoutput=1

\documentclass[11pt]{article}

\usepackage[final]{acl}

\usepackage{times}
\usepackage{latexsym}
\usepackage{multirow}
\usepackage{booktabs}
\usepackage[T1]{fontenc}

\usepackage[utf8]{inputenc}

\usepackage{microtype}
\usepackage{graphicx}
\usepackage{subcaption}
\usepackage{amsmath}
\usepackage{cleveref}
\usepackage{tcolorbox}

\usepackage{amssymb}
\usepackage{xcolor}
\usepackage{pifont}
\definecolor{my_green}{RGB}{51,102,0}
\definecolor{my_red}{RGB}{204, 0, 0}
\newcommand{\cmark}{\textcolor{my_green}{\ding{51}}} 
\newcommand{\xmark}{\textcolor{my_red}{\ding{55}}} 

\usepackage{inconsolata}

\title{Multimodal ArXiv: A Dataset for Improving Scientific Comprehension of Large Vision-Language Models}

\author{
Lei Li\thanks{Equal Contribution.}$^\dagger$,  Yuqi Wang$^*$$^\dagger$, Runxin Xu$^\ddag$, Peiyi Wang$^\ddag$\\\textbf{Xiachong Feng$^\dagger$, Lingpeng Kong$^\dagger$, Qi Liu$^\dagger$}\\ 
$^\dagger$The University of Hong Kong \\
$^\ddagger$Peking University\\
 \texttt{\{nlp.lilei, runxinxu, wangpeiyi9979, xiachongfeng1996\}@gmail.com} \\ 
 \texttt{wangyuqi@connect.hku.hk} \quad \texttt{\{lpk, liuqi\}@cs.hku.hk}\\
  }
\def\DatasetName{ArXivCap\ }

\begin{document}
\maketitle
\input{00_abstract}

\input{01_intro}

\input{02_related}

\input{dataset}

\input{04_experiment}

\input{10_conclusion}

\input{ack}

\bibliography{11_references}
\input{12_appendix}
\end{document}

%% file: 00_abstract.tex
\begin{abstract}

Large vision-language models (LVLMs) excel across diverse tasks involving concrete images from natural scenes. 
However, their ability to interpret abstract figures, such as geometry shapes and scientific plots, remains limited due to a scarcity of training datasets in scientific domains.
To fill this gap, we introduce Multimodal ArXiv, consisting of ArXivCap and ArXivQA, for enhancing LVLMs scientific comprehension.
ArXivCap is a figure-caption dataset comprising 6.4M images and 3.9M captions, sourced from 572K ArXiv papers spanning various scientific domains.
Drawing from ArXivCap, we introduce ArXivQA, a question-answering dataset generated by prompting GPT-4V based on scientific figures. 
ArXivQA greatly enhances open-sourced LVLMs' mathematical reasoning capabilities, achieving a 10.4\% absolute accuracy gain on a multimodal mathematical reasoning benchmark.
Furthermore, employing ArXivCap, we devise four vision-to-text tasks for benchmarking LVLMs.
Evaluation results with state-of-the-art LVLMs underscore their struggle with the nuanced semantics of academic figures, while domain-specific training yields substantial performance gains.
Our error analysis uncovers misinterpretations of visual context, recognition errors, and the production of overly simplified captions by current LVLMs, shedding light on future improvements.\footnote{Datasets and models are released at our project page: \url{https://mm-arxiv.github.io}.}

\end{abstract}

%% file: 01_intro.tex
\section{Introduction}
\label{sec:intro}

Large vision-language models (LVLMs), which integrate large language models (LLMs)~\citep{gpt3,touvron2023llama} with pre-trained vision encoders through cross-modal alignment training~\citep{madureira-2021-flamingos,liu2023llava,li2023m3it}, 
have demonstrated remarkable perceptual and cognitive capabilities in processing concrete images from everyday scenes~\citep{gpt4v,fu2023mme,yang2023dawn,reka2024core}.
However, recent studies have shown that open-source LVLMs struggle to understand abstract figures, such as geometric shapes in multimodal mathematical reasoning~\citep{mathvista,zhang2024mathverse} and scientific plots~\citep{mmmu}. 
The inadequacy of training datasets in scientific domains that involve complex reasoning with abstract figures is the main underlying cause.

To address this, we construct Multimodal ArXiv by utilizing the rich resources in preprints hosted on arXiv to improve the ability to understand scientific literature in LVLMs.
We first curate ArXivCap, a
diverse scientific figure-caption dataset.
In contrast to previous scientific figure datasets, which consist of synthesized figures~\citep{chen2020figcap}
or are restricted to simple captioning scenarios in the computer science domain~\citep{hsu-etal-2021-scicap-generating}, our dataset is composed of figures extracted from academic papers across a range of domains.
ArXivCap has 6.4M images and 3.9M captions from 572K papers. 
We also keep the subfigure structure, and titles of original papers, thereby supporting diverse evaluation tasks.
We further instruct GPT-4V to generate 100K multiple-choice question-answering~(QA) pairs for the figures in ArXivCap.
The resulting ArXivQA dataset could naturally serve as a pivotal resource for improving the scientific reasoning abilities of LVLMs.

We validate the effectiveness of our Multimodal ArXiv dataset from two dimensions: reasoning ability measured by QA accuracy and generation performance through novel vision-to-text tasks. 
Our experiments demonstrate that ArXivQA brings a significant 10.4\% absolute accuracy boost for Qwen-VL-Chat~\citep{Qwen-VL}, on the MathVista~\citep{mathvista}, a challenging benchmark for multimodal mathematical reasoning. 
Additionally, detailed analysis uncovers the relationship between paper domains and fine-grained task performance.
Moreover, using ArXivCap, we define four generation tasks of varying complexity to benchmark the ability of LVLMs to comprehend scientific plots:
(1) captioning a single academic figure, (2) generating overall summaries for multiple sub-figures, 
(3) in-context figure captioning given previous figure-caption pairs, and (4) generating paper titles from figure-caption pairs. 
We examine various LVLMs, including open-source models as well as proprietary models including GPT-4V~\citep{gpt4v} and Gemini 1.0 Pro Vision~\citep{team2023gemini}.
Evaluation results reveal that despite that current LVLMs still face challenges generating faithful captions for scientific figures, in-domain training on our dataset yields substantial performance improvements across all four tasks.
Manual error analysis underscores that LVLMs still suffer from misinterpretation of the visual context, recognition errors, and overly simplified captions, paving the way for future studies.




%% file: 02_related.tex
\section{Related Work}
\label{sec:related}
Recent advancements in LVLMs have seen notable progress in model architecture, training paradigms, and dataset creation~\citep{zhang2024mm}.

\paragraph{Model Architecture} LVLMs typically comprise three core modules: (i) a vision encoder for image feature extraction, (ii) a modality alignment module to integrate visual features into the language model embedding space, and (iii) an LLM backbone for decoding multimodal context. 
CLIP~\citep{radford2021clip} is widely used for image encoding, while LLaMA~\citep{touvron2023llama} and Vicuna~\citep{vicuna2023} serve as popular choices for LLMs. 
The alignment module varies from simple linear projections~\citep{liu2023llava,zhu2023minigpt4} to more complex architectures like gated cross-attention layers~ substantiated by Flamingo and IDEFICS~\citep{Alayrac2022FlamingoAV,awadalla2023openflamingo}. Innovations such as the Q-Former module in BLIP2~\citep{li2023blip2} and instruction integration in InstructBLIP~\citep{dai2023instructblip} further enhance alignment capabilities. 
Additionally, Fuyu-8B~\citep{fuyu-8b} introduces a novel framework mapping raw image pixels directly to the LLM embedding space.

\paragraph{Training Paradigms} Regarding the training recipes, PaLI-X~\citep{Chen2023PaLIX} investigates the scaling effects of both vision encoders and language models, highlighting the advantages of scaling both components. Qwen-VL~\citep{Qwen-VL} increases input image resolution and explores different module unfreezing strategies. Alignment methodologies such as RLHF training~\citep{ouyang2022instructgpt}, e.g., LLaVA-RLHF~\citep{2023llavarlhf}, and preference optimization through AI feedback~\citep{2023vlfeedback}
demonstrate effectiveness in aligning LVLMs with human preferences.

\paragraph{Dataset Curation} Dataset quality significantly impacts LVLM performance. Modality alignment training often utilizes web-scale image-caption pairs such as Laion-400M~\citep{laion400m}, with recent studies favoring cleaned captions~\citep{sharegpt4v,capsfusion}. Instruction fine-tuning~(IFT) helps LVLMs respond according to user queries, triggering the exploration of high-quality IFT datasets. 
Efforts include multimodal instruction collections such as MultiInstruct~\citep{multiinstruct} and M$^3$IT~\citep{li2023m3it}, dialog-style datasets such as LLaVA~\citep{liu2023llava} and domain-specific datasets for medical~\citep{med-llava} and text-rich images~\citep{llavar}.
In the scientific domain, FigCAP~\citep{chen2019figcap} and FigureQA~\citep{kahou2017figureqa} are created based on synthetic figures.
DVQA~\citep{kafle2018dvqa} creates heuristic-based questions for bar charts only.
SciCap~\citep{hsu-etal-2021-scicap-generating}, SciCap+~\citep{Yang2023SciCap+}, and M-Paper~\citep{hu2023mplugpaperowl} collect figure-caption pairs from specific domains such as computer science.
Compared with these datasets, our ArXivCap is sourced from diverse scientific domains with a much larger scale, enabling more comprehensive improvements and evaluations. Besides, we employ GPT-4V for creating ArXivQA with challenging questions, showcasing its effectiveness in boosting the mathematical reasoning ability of LVLMs.


%% file: dataset.tex
\section{Multimodal ArXiv}
This section presents a detailed construction process of our Multimodal ArXiv dataset, consisting of two sets: ArXivCap~(\S\ref{subsec:ArXivcap}) and ArXivQA~(\S\ref{subsec:arxiv_qa}).


\input{figs/process-overview}

\subsection{ArXivCap}
\label{subsec:ArXivcap}

\paragraph{Construction Process} We outline the creation process of ArXivCap below and Figure~\ref{fig:dataset-curation-process} gives an overview.

\noindent\emph{Paper Filtering with Publication Type:}
\DatasetName is extracted from ArXiv~\cite{clement2019use}, which is under CC-0 licence for modification and distribution. 
The raw files of papers posted on ArXiv tar files before June 2023 are downloaded. 
To ensure the quality of our dataset, we employ a rigorous selection process to filter potentially low-quality papers that might influence the figure-caption pair quality. 
Firstly, we retrieve meta-information for papers from Semantic Scholar~\cite{kinney2023semantic}, which contains the publication record for each paper. 
Papers with publication types \texttt{JournalArticle},  \texttt{Conference}, or \texttt{Review} are kept as we assume the peer-review process could ensure the overall figure-caption quality is satisfactory.
We further exclude papers with titles exceeding 100 words or abstracts longer than 300 words, in alignment with common submission requirements.



\begin{figure}[t!]
    \centering
    \includegraphics[width=0.9\linewidth]{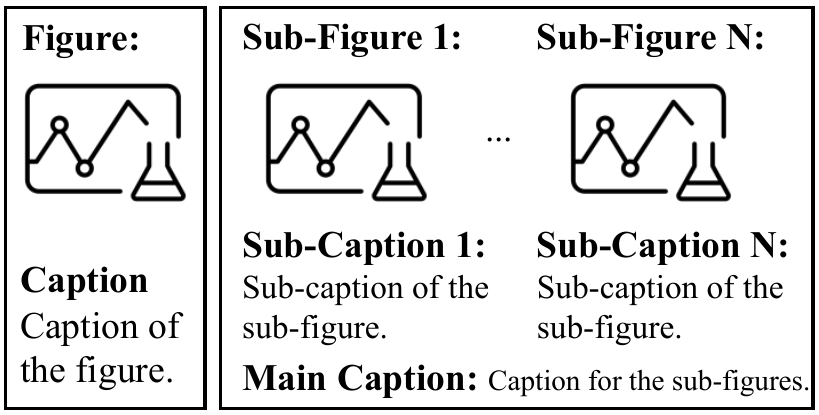}
    \caption{Illustration of two types of figure-caption pairs. (Left) Single-Figure pair. (Right) Multiple-Figure caption pair has multiple sub-figures with corresponding sub-captions and an overall main caption. }
    \label{fig:chunk_example}
\end{figure}

\noindent\emph{Figure-Caption Pair Extraction:}
Images and captions are extracted from the original LaTeX files by matching the syntax. 
We further use a robust tool ImageMagisk~\cite{imagemagick} to convert images into JPEG format for easy processing.
The extracted images and captions are stored in a designed chunk structure,
which consists of either a single figure-caption pair or multiple figures with their respective sub-captions and a main caption for the overall description.
This format is more consistent with the layout of academic papers, and
Figure~\ref{fig:chunk_example} illustrates the chunk structure.


\noindent\emph{Caption Cleaning and Image Filtering:}
After a manual inspection of the initially collected dataset, we design several transformations to clean the captions and filter the images.

\noindent\emph{Caption Cleaning}: (i) Chunks with captions shorter than 5 words are removed; (ii) For captions with LaTeX expressions such as math formulas and references, we apply the \texttt{pylatexenc}\footnote{https://github.com/phfaist/pylatexenc} to transform the LaTeX to text with math formulas retained, citations to a special symbol \texttt{<cit.>}, references to \texttt{<ref>}. An illustration of caption cleaning can be found in Appendix~\ref{apx:caption_clean}.

\noindent\emph{Image Filtering}: We remove images that are deemed to be problematic according to the following rules:
(i) Images with an aspect ratio larger than 100; (ii) Images with the shortest edge shorter than 224 pixels; and (iii) Images with pixel numbers larger than the decompression bombs threshold.

After these processes, 100 pairs are sampled to perform an additional manual inspection, where we found all of these pairs contained clear images and correct caption descriptions. We provide visualized figure-caption pairs in Appendix~\ref{apx:case_illustrations}.
\paragraph{Statistics of ArXivCap}

\input{tables/detail_statistic_title_abstract_image_caption}

\input{tables/compare}

Table~\ref{tab:detail_statistic_title_abstract_image_caption} lists the dataset statistics. 
ArXivCap consists of 572K papers, containing 6.4M high-quality images in total with 193M words. 
A word cloud illustration of captions can be found in the Appendix~\ref{apx:caption_word_cloud}.
Figure~\ref{fig:domain-distribution} demonstrates the paper domain distribution extracted from ArXiv, where we find that our ArXivCap covers 32 domains, such as computer science, mathematics, physics, and economics.
As shown in Table~\ref{tab:dataset_comparison_1}, compared with previous scientific figure datasets, our ArXivCap is the largest figure-caption dataset collected from real papers and covers a wide range of scientific domains, serving as a valuable resource for improving and benchmarking LVLMs.





\input{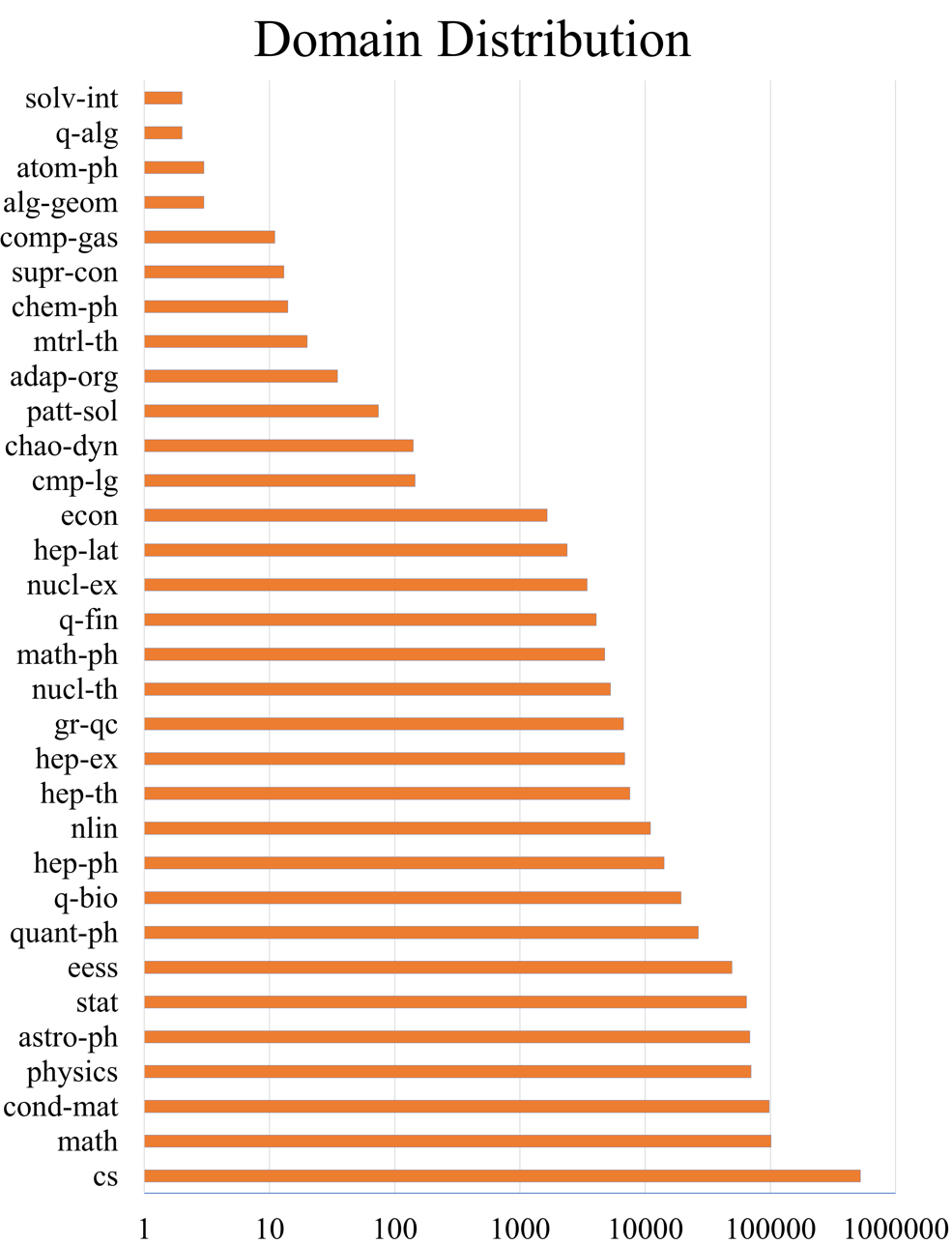}

\subsection{ArXivQA}
\label{subsec:arxiv_qa}
As our ArXivCap contains diverse images from scientific domains, we assume that learning to answer questions about these figures could boost scientific reasoning ability. Following the successful practice of LLaVA~\citep{liu2023llava}, we adopt GPT-4V to generate instruction-tuning datasets for generating the QA pairs based on the figures extracted from scientific papers.
Specifically, we design a prompting template to query GPT-4V for generating QA pairs based on 35K images randomly sampled from our ArXivCap.
Table~\ref{tab:prompt_for_ArXivqa} in Appendix~\ref{apx:prompt_template} provides the template we used for the prompt.
The generated pairs are parsed according to the format requirement and we discard the samples without options and rationales.
There are 100K QA pairs after filtering the invalid samples. 
The dataset comprises questions with an average word count of 16.98 for the question text.
On average, there are 4.20 options per question and the average length of the text for a single option is 7.59 words.
Appendix~\ref{apx:case_illustrations} provides samples from the ArXivQA dataset.

As a preliminary study, we sample 1,000 samples from ArXivQA and prompt open-sourced LVLMs to predict answers given the questions and options. 
A simple prompt is designed to employ GPT-4 for extracting the answer label from the model generations.
For human performance, we ask four authors to perform predictions on a 100-sample subset (where 17 samples are from the CS domain). Each of them is asked to answer 50 samples and the accuracy scores are obtained by averaging two annotators.
As shown in Table~\ref{tab:arxiv_qa_acc}, most models struggle to perform satisfactorily on the ArXivQA dataset, falling far behind human performance. This verifies our premise that current open-sourced LVLMs fail to understand scientific figures. We also notice that simply increasing the model scale from 7B (LLaVa-1.5-7B) to 13B (LLaVa-1.5-13B) does not yield a significant boost, which indicates that the ability for multimodal mathematical reasoning cannot be simply acquired from the LLM-side only.

\input{tables/arxivqa_acc}

%% file: figs/process-overview.tex
\begin{figure*}[tp]
    \centering
    \includegraphics[width=0.95\linewidth]{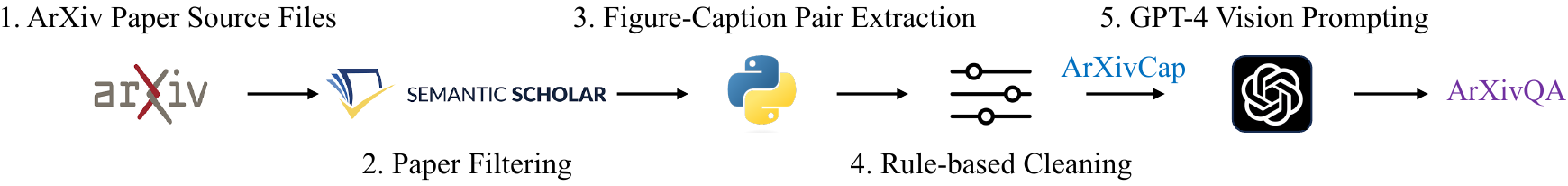}
    \caption{Overview of our dataset curation process. Starting from the ArXiv paper source files, we ensure the paper quality by selecting papers according to publication records. Figure and caption pairs are extracted and then cleaned according to manually designed rules. ArXivQA is generated by prompting GPT-4V with a curated template.}
    \label{fig:dataset-curation-process}
\end{figure*}

%% file: tables/detail_statistic_title_abstract_image_caption.tex
    
    
         
                 
        
        



\begin{table}[t!]
    \centering
    \resizebox{\linewidth}{!}{
    \begin{tabular}{@{}lccc@{}}
         \toprule
        Field & Number &  Average Len.  & Quartile of Len. \\
         \midrule
         
        Title & 572K &  10.4   & (8, 10, 12) \\ 
        Abstract & 572K &  167.6  & (126, 165, 207) \\
        \midrule 
        Main Caption & 3.9M  &  47.6 & (15, 35, 65) \\
        Subcaption & 1.0M  & 4.8 &  (2, 3, 5) \\  
        Chunk Caption & 3.9M &  48.8  & (16, 36, 67) \\
        \midrule 
        Images & 6.4M  & N / A  & N / A \\
         \bottomrule
    \end{tabular}}
    \caption{Word count statistics for title, abstract, captions, and Image number. 
    Chunk caption refers to the combination of subcaptions and the main caption for a multiple-figure case.}
    \label{tab:detail_statistic_title_abstract_image_caption}
\end{table}

%% file: tables/compare.tex
\begin{table*}[ht!]
    \centering
\resizebox{\textwidth}{!}{
    \begin{tabular}{@{}lccccc@{}}
         \toprule
        Dataset & Image Number & Paper Number & Image Category & Domain & Real Data \\
         \midrule
                 FigCAP~\citep{chen2020figcap} & 219K & N / A & Bar, Line and Pie Charts & N / A & \xmark \\
        
         SciCap~\citep{Yang2023SciCap+} & 2.1M & 295K & Open-Category & Computer Science and Machine Learning & \cmark \\

         M-Paper~\citep{hu2023mplugpaperowl} & 350K & 48K & Open-Category & Mainly "Deep Learning" & \cmark \\   
        
         ArXivCap (Ours) & 6.4M & 572K & Open-Category & Open-Domain & \cmark \\
 
        \midrule 
FigureQA~\citep{kahou2017figureqa}  & 140K & N / A & Bar, Line and Pie Charts & N / A & \xmark \\
                 
        DVQA~\citep{kafle2018dvqa} & 300K & N / A & Bar Charts & N / A& \xmark \\
         ArXivQA (Ours) &  32K & 16.6K & Open-Category & Open-Domain & \cmark \\
         \bottomrule
    \end{tabular}}
    \caption{Comparison with previous scientific figure datasets. Our ArXivCap is the largest captioning dataset and our ArXivQA is the only QA dataset that covers a wide range of domains from real papers. }
    \label{tab:dataset_comparison_1}
\end{table*}


%% file: figs/domain-distribution.tex
\begin{figure}[t!]
    \centering
    \includegraphics[width=0.9\linewidth]{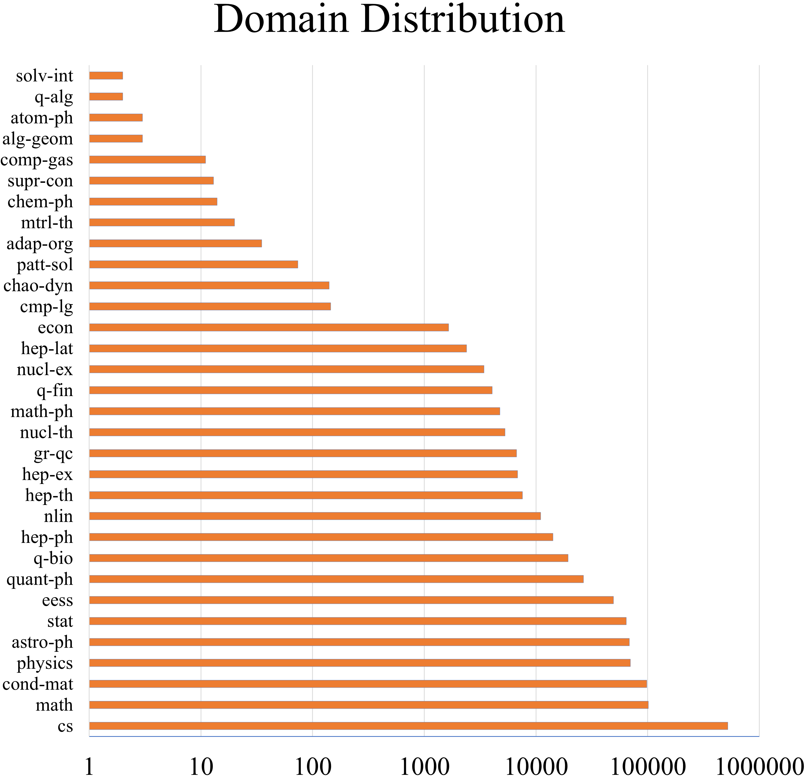}
    \caption{Paper domain distribution of ArXivCap. See Table~\ref{tab:domain-full-name} in Appendix~\ref{apx:arxiv_cap} for the full name of each domain.}
    \label{fig:domain-distribution}
\end{figure}

%% file: tables/arxivqa_acc.tex
\begin{table}[t!]
    \centering
    \small 
    \begin{tabular}{l|c}
        \toprule
        Model & Accuracy \\
        \midrule
        InstructBLIP-Vicuna7B & 7.0\% \\
        LLaVA-1.5-7B & 44.2\% \\
        LLaVA-1.5-13B & 46.8\% \\
        OpenFlamingo-9B & 9.9\% \\
        IDEFICS-Instruct-9B & 34.5\% \\
        Qwen-VL-Chat & 46.6\% \\
        \midrule 
        \textcolor{gray}{Human (100-sample Subset)} & \textcolor{gray}{80.0\%} \\
        \textcolor{gray}{Human (CS subset)} & \textcolor{gray}{88.2\%} \\
        \bottomrule
    \end{tabular}
    \caption{Evaluation results on the sampled 1,000 ArXivQA samples. }
    \label{tab:arxiv_qa_acc}
\end{table}

%% file: 04_experiment.tex
\input{tables/ArXivqa}

\section{Experiments}

We conduct experiments to (i) validate the effectiveness of ArXivQA for boosting multimodal scientific reasoning for open-source LVLMs~(\S\ref{subsec:math_reasoning}) and (ii) benchmark LVLMs capability to comprehend scientific figures with ArXivCap~(\S\ref{subsec:exp_arxivcap}).

\input{math_reasoning}

\input{03_method}

\subsubsection{Experimental Settings}
\label{subsubsec:arxiv_cap_setting}

\paragraph{Dataset}
We divide ArXivCap into training and test sets with a 9:1 ratio for evaluation. The test set includes:
161.3K samples for single-figure captioning,
12.8K samples for multiple-figure captioning,
57.2K samples for contextualized captioning, and
57.2K samples for title generation.

\paragraph{Evaluated Models}
We select various LVLMs covering different architectures. 
(1) LVLMs designed for dealing with a single image, BLIP2-OPT-6.7B~\citep{li2023blip2}
, InstructBLIP-Vicuna7B~\citep{dai2023instructblip}, 
LLaVA-1.5-7B/13B~\citep{liu2023llava15}. Due to the ability limitation, we only benchmark these models on the single image captioning task;
(2) LVLMs capable of handling interleaved text-image inputs, such as OpenFlamingo-9B~\citep{Alayrac2022FlamingoAV,awadalla2023openflamingo}, IDEFICS-Instruct-9B~\citep{laurencon2023obelics}, Qwen-VL-Chat-7B~\citep{Qwen-VL}. These models are evaluated on all the tasks we proposed;
(3) Proprietary models such as Gemini 1.0 Pro Vision and GPT-4V.
Due to the large scale of our test set, we randomly sample a subset consisting of 500 instances for evaluating these two models to reduce costs, with corresponding scores colored in \textcolor{gray}{grey}. Details of evaluated models and the task prompts used are provided in Appendix~\ref{apx:evaluation_details}.

\paragraph{Training Settings} 
To investigate whether in-domain training can enhance the model's capabilities, we train the Qwen-VL-Chat-7B on ArXivCap using the same setting as in \S\ref{subsubsec:exp_setting}. To fit the input length limit, we set the maximum number of figures per sample to four. The training process takes 70 hours with 8 NVIDIA A100s.

\paragraph{Metrics}BLEU-2~\citep{bleu}, ROUGE-L~\citep{lin-2004-rouge} and BERT-Score~\citep{bertscore} are adopted as the automatic evaluation metrics.
We also explore using GPT-4 to assist in caption evaluation. Our findings in Appendix \ref{apx:gpt4_eval} indicate that ROUGE-L and BLEU-2 scores are highly correlated with GPT-4's annotations. We primarily use these three metrics due to their convenience. A manual error analysis is conducted to supplement the automatic metrics~(\S\ref{subsubsec:manual_eval}).


\begin{figure}[t!]
    \centering
    \includegraphics[width=\linewidth]{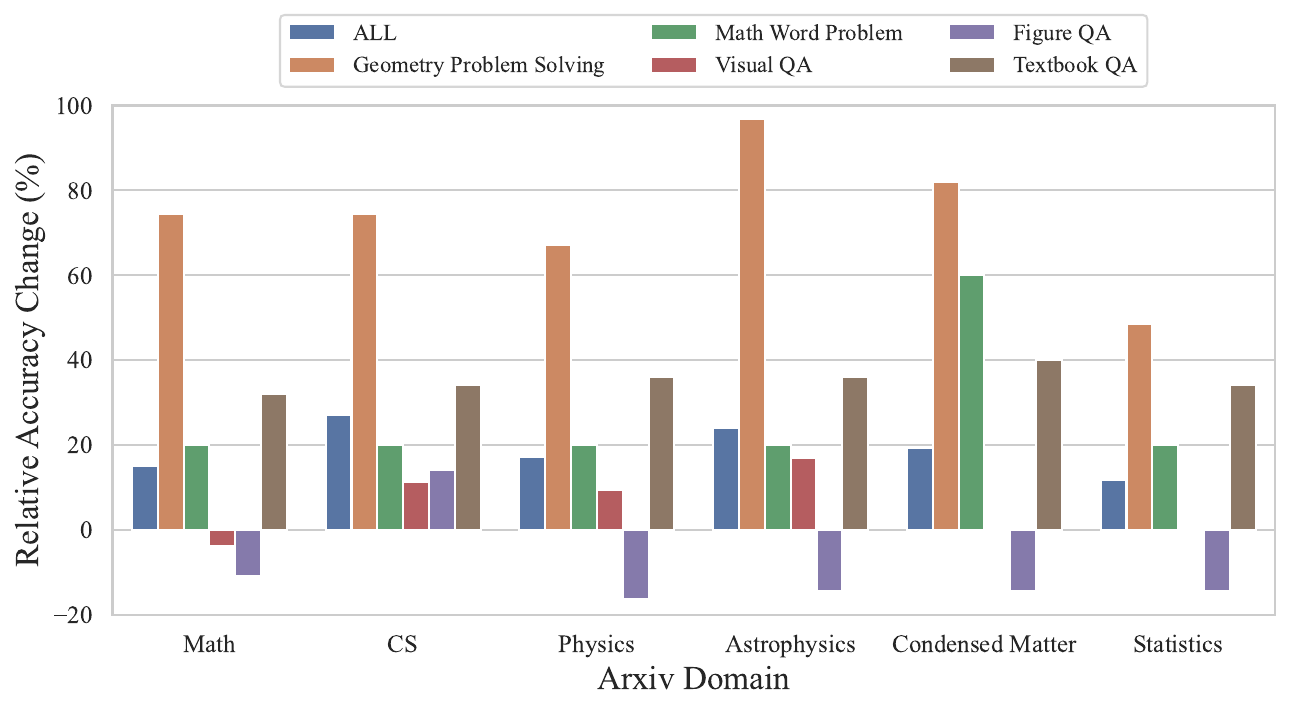}
    \caption{Relative accuracy changes brought by the training on different domain ArXivQA samples.}
    \label{fig:domain_analysis}
\end{figure}
\input{tables/single_cap}
\input{tables/cap_with_title_abstract}
\input{tables/main_ret}
\subsubsection{Results}
\paragraph{Results of Single-Figure Captioning}
The evaluation results for the single-figure captioning task are presented in Table \ref{tab:single_cap_ret}. Despite achieving near-perfect performance on conventional image captioning tasks like MSCOCO \citep{lin2014mscoco}, open-source LVLMs, such as LLaVA models, face challenges when applied to academic figures. 
For closed models, GPT-4V performs comparably with Gemini 1.0 Pro Vision.
Furthermore, continuous training on our dataset yields a significant performance boost for this task. For instance, fine-tuning results in a notable increase in the BLEU-2 score from 4.4 to 8.9, indicating a promising avenue for enhancing academic figure comprehension through domain-specific training.
We also investigate whether providing additional context information, such as the paper title and abstract, could help models generate better figure captions. As shown in Table \ref{tab:cap_with_meta_ret}, adding the title is beneficial evidenced by the boosted scores, while providing abstracts brings negligible gains.


\paragraph{Results of Multiple-Figure Captioning}
As shown in the first block of Table \ref{tab:ret_three_tasks}, similar to single-figure captioning, multiple-image captioning poses a challenge for current open-source LVLMs. For instance, Qwen-VL-Chat achieves only a 3.0 BLEU-2 and a 7.2 ROUGE-L score on this task, considerably lower than its performance in single-figure captioning. In contrast, GPT-4V consistently demonstrates proficiency in both tasks, suggesting a balanced ability to capture semantics across multiple images. Notably, training on our ArXivCap dataset yields more pronounced improvements for this task, culminating in Qwen-VL-Chat even surpassing the performance of the GPT-4V model. This enhancement underscores the pivotal role of our dataset in facilitating LVLMs to enhance reasoning capabilities over multiple images, leading to more effective summarization of scientific figures.

\paragraph{Results of Contextualized Captioning}
In the middle block of Table~\ref{tab:ret_three_tasks}, we find that IDEFICS-Instruct-9B achieves the best performance on this task.
This achievement is largely attributed to its remarkable proficiency in leveraging contextual cues, stemming from its extensive pre-training involving interleaved image-text pairs~\citep{laurencon2023obelics}. 
Interestingly, fine-tuning on ArXivCap results in marginal performance declines across all metrics, with GPT-4V achieving the lowest scores as well. 
This phenomenon can be attributed to the tendency of sequential captions to exhibit similar patterns, thereby favoring models that effectively leverage contextual cues.
We perform two more challenging evaluations by (i) providing context pairs from another paper and (ii) randomly shuffling the order of figure-caption pairs in the context.
As shown in Table~\ref{tab:order_analysis}, 
the performance with random contexts degrades significantly, validating our previous hypothesis.
Instead, the fine-tuned model demonstrates more robust captioning results under these settings, evidenced by the slight 8\% drop on ROUGE-L compared to the 31\% of the original model with shuffled context orders.

\begin{table}[t!]
    \centering
    \resizebox{\linewidth}{!}{
    \begin{tabular}{l|cc}
    \toprule
       Model  &   BLEU-2 ($\Delta\downarrow$) & ROUGE-L ($\Delta\downarrow$)  \\
    \midrule
        Qwen-VL-Chat-7B & 17.0 & 22.1\\ 
       \quad + random contexts & 	5.7 (66.5\%)&  13.0 (38.1\%)\\ 
       \quad  + shuffle order& 12.0 (29.4\%) &15.1 (31.7\%) \\ 
       
    \midrule 
      Qwen-VL-Chat-7B$_\text{ArXivCap}$ & 16.1 & 21.2\\ 
       \quad + random contexts	& 7.5 (53.4\%)	& 14.3 (32.5\%)\\ 
        \quad + shuffle order & 14.1 (12.4\%) & 19.5 (8.0\%) \\
    
    \bottomrule
    \end{tabular}}
    \caption{Contextualized captioning performance is influenced by the order. After tuning on the ArXivCap, the model is more robust to the order of the history captions.}
    \label{tab:order_analysis}
\end{table}
\paragraph{Results of Title Generation}
The results are presented in the last block of Table~\ref{tab:ret_three_tasks}.
Notably, the title generation task poses a formidable challenge, evident in the significantly lower overall BLEU-2 score compared to the captioning tasks. This suggests the inherent difficulty in generating precise predictions for paper titles.
A contrasting picture emerges when considering the ROUGE-L and BERT-Score metrics, which either closely align or surpass the performance on captioning tasks. This underscores the model's proficiency in producing semantic-related results given the presented figures. 
Consistent with the previous two tasks,
fine-tuning the model on our dataset yields substantial enhancements for the title generation task. The BLEU-2 score jumps impressively from 2.6 to 6.7, while the ROUGE-L score sees a commendable increase from 15.8 to 23.5. 
These findings highlight the challenge of title generation for current LVLMs and the effectiveness of our dataset in improving the model's capability to generate accurate titles.






\subsection{Analysis}

\begin{figure}
    \centering
\includegraphics[width=0.8\linewidth]{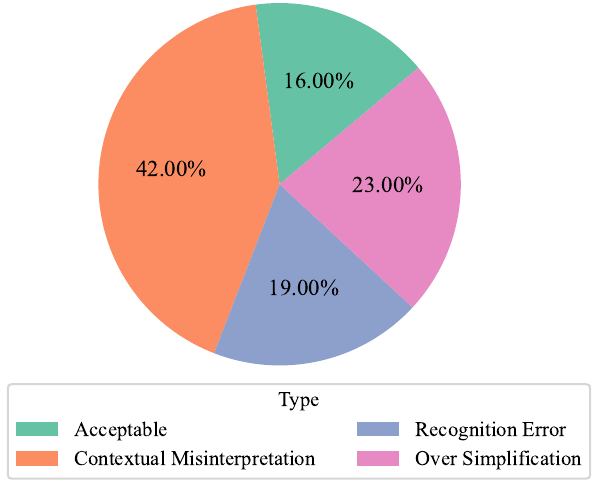}
    \caption{Manual analysis of the generated captions. }
    \label{fig:error_pie_chart}
\end{figure}

\begin{figure*}
    \centering
    \includegraphics[width=0.62\linewidth]{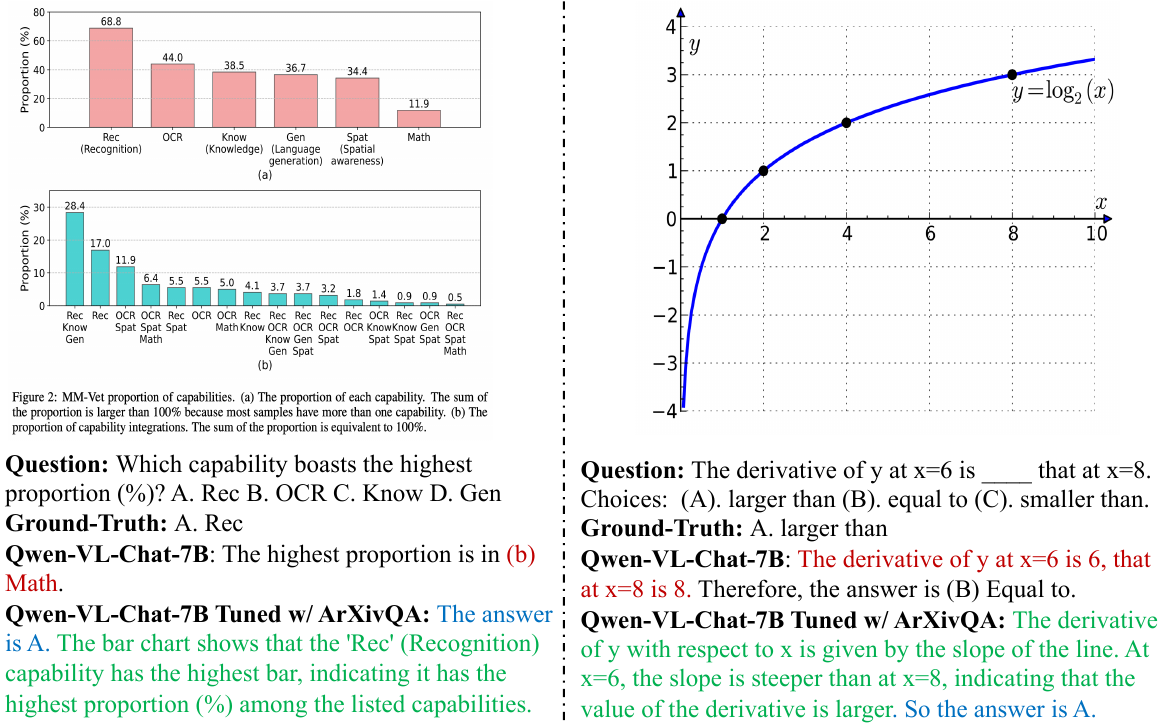}
    \caption{ArXivQA enables the model not only to answer questions related to scientific figures in papers (left) but also to improve mathematical understanding ability (right). The model not only selects correct options but also gives reasonable rationale.}
    \label{fig:math_case_study}
\end{figure*}

\paragraph{Manual Evaluation of Generated Captions} 
\label{subsubsec:manual_eval}
We conduct a manual inspection for single-figure captioning results. To ensure a more informed evaluation, we focus on a paper from the CS domain, leveraging our domain knowledge to assess caption quality better. 
The quality of generated captions is assessed by scrutinizing the figure, the ground-truth caption, the paper title, and the abstract.
We categorize captions into the following quality types according to our preliminary inspection: (1) \emph{Acceptable}, where captions accurately encapsulate the scientific figure's essence, aligning with the intended information of the ground-truth; (2) \emph{Over Simplification}, instances where the model oversimplifies content, offering a broad overview while neglecting specific details and nuances present in the ground truth; (3) \emph{Recognition Error}, where the model inaccurately recognizes and describes key visual and textual elements in the scientific figure, such as colors, numerical values, or textual context; and (4) \emph{Contextual Misinterpretation}, where the model misinterprets the specific context of the scientific figure, resulting in captions relevant in a generic sense but inaccurate for the given figure. Visualized generated captions of different types are shown in Figure~\ref{fig:caption_type} of Appendix~\ref{apx:caption_type}.
The results of 100 manually examined captions are depicted in Figure~\ref{fig:error_pie_chart}, revealing that only 16\% of captions are deemed acceptable when compared to human-written ones. 
Among unsatisfactory captions, contextual misinterpretation emerges as the dominant issue, suggesting a need for incorporating more contextual information as suggested in Table~\ref{tab:cap_with_meta_ret}. Oversimplification is another concern, with generic captions identified. Additionally, 23\% and 19\% of examined captions suffer from the oversimplification issue and recognition errors in reported numbers/texts in the caption, respectively. The former is attributed to the highly frequent simple caption in the training dataset and the latter issue could be addressed through potential integration with OCR results.
Our manual evaluation suggests future efforts may benefit from incorporating additional context clues, such as paper metadata, improving the model's fundamental perception abilities, and utilizing external information.


\paragraph{Case Study of MathVista}
We conduct case studies to illuminate the tuning effects facilitated by our ArXivQA dataset. In the left segment of Figure~\ref{fig:math_case_study}, ArXivQA helps the model accurately answer a question related to the presented bar plot. 
The right part in Figure~\ref{fig:math_case_study} demonstrates that ArXivQA can enhance algebraic reasoning abilities. Here, a question involving the derivative of a function is correctly answered, accompanied by a lucid reasoning rationale.
Figure~\ref{fig:geometry_fail} in Appendix~\ref{apx:failure_mathvista} highlights a challenging geometry problem where both models generate hallucinated outputs. 
These illustrative cases collectively affirm the efficacy of our dataset.


%% file: tables/arxivqa.tex
\begin{table*}[t!]
    \centering
    \resizebox{\linewidth}{!}{
    \begin{tabular}{@{}l|ccccc|c@{}}
    \toprule 
     Model    &   Figure QA & Geometry Problem Solving &  Math Word Problem  & Textbook QA  & Visual  QA & ALL \\
     \midrule
     IDEFICS-Instruct-9B$^\dagger$ & 41.4  & 22.0 & 18.2 &  34.6  & 44.6 & 33.7 \\ 
     InstructBLIP-Vicuna13B$^\dagger$ & 41.4  & 19.9 & \underline{45.5} & 45.8  & 57.6 & 39.3 \\ 
     LLaVa-v1.5-13B$^\dagger$ & 44.0  & 26.7 & 40.9 &  45.8 &  44.6 & 39.3 \\

     \midrule 
   Qwen-VL-Chat-7B & \underline{48.3} & 19.1 &22.7 &46.7& 57.6& 40.0\\ 
   Qwen-VL-Chat-7B$_\text{ArXivCap}$  & 39.7&  19.8& 27.2 & 39.7 & 52.1 & 36.2\\ 
   Qwen-VL-Chat-7B$_\text{ArXivQA}$ & {44.8}& 34.0& 27.3 & \underline{70.0} & \underline{64.1}  & {50.2}\\ 
   Qwen-VL-Chat-7B$_\text{ArXivCap + ArXivQA}$ & 44.0& 37.6 & 27.3 & 68.2 & 63.0 & \underline{50.4} \\ 
    \midrule 
    Bard$^\dagger$ &  38.8 & \underline{51.1} & 27.3& 64.5 & 51.1  &50.0 \\ 
    GPT-4V$^\dagger$ & \textbf{52.6} & \textbf{51.8} & \textbf{54.5} & \textbf{83.2}& \textbf{66.3}   &  \textbf{61.9} \\ 
    \bottomrule
    \end{tabular}}
    \caption{Evaluation on MathVista dataset. ArXivCap and ArXivQA together enhance Qwen-VL-Chat's overall performance, surpassing that of the commercial model Bard. $^\dagger$ denotes results based on the original predictions from \citet{mathvista}. The best results are highlighted in \textbf{bold}, while the second-best scores are marked with \underline{underline}.}
    \label{tab:mathvista_ret}
\end{table*}

%% file: math_reasoning.tex
\subsection{Boosting LVLMs with ArXivQA}
\label{subsec:math_reasoning}

\subsubsection{Experimental Settings}
\label{subsubsec:exp_setting}
We adopt Qwen-VL-Chat-7B~\citep{Qwen-VL}
as the backbone due to its support for interleaved image-text input formats and high-resolution images.
We fine-tune it on our ArXivCap (Qwen-VL-Chat-7B$_\text{ArXivCap}$), ArXivQA (Qwen-VL-Chat-7B$_\text{ArXivQA}$) and combination of these two datasets (Qwen-VL-Chat-7B$_\text{ArXivCap + ArXivQA}$) for three epochs with a learning rate of 1e-5 following the original paper.
We combine the answer and the rationale in ArXivQA to form the target output during training. 
Models are evaluated on MathVista~\citep{mathvista}, a benchmark that requires fine-grained, deep visual understanding and compositional reasoning. MathVista contains 6,141 examples, consisting of five multimodal tasks Figure QA, Geometry Problem Solving, Math word problem, Text Book QA, and Visual QA. 
We select 478 multiple-choice questions in the \texttt{testmini} split to avoid the inconsistency of answer parsing. We compute the accuracy scores and adopt the provided prediction files for calculating the baseline performance.

\subsubsection{Results}

As shown in Table~\ref{tab:mathvista_ret},
fine-tuning on our Multimodal ArXiv, especially on the ArXivQA dataset, consistently boosts the performance, helping the open-sourced Qwen-VL-Chat achieve a comparable overall MathVista reasoning performance.
Due to the wide coverage of the scientific figures, the performance gain mainly comes from significantly improved Geometry Problem Solving, Textbook QA, and Visual QA tasks. For example, after fine-tuning on the ArXivQA dataset, the accuracy is increased from 19.1\% to 34.0\% and from 46.7\% to 70.0\% on Geometry Problem Solving and Textbook QA tasks, respectively.
The improvement on Math Word Problem is marginal, where we think the domain-specific data augmentation can be further explored with a curated filtering dataset on our dataset~\citep{gao2023gllava}.
On the contrary,
the accuracy of Figure QA deteriorates slightly compared with the original backbone model, which we attribute to the fact that most of the plots in the Figure QA evaluation are sampled from synthesized datasets such as DVQA~\citep{kafle2018dvqa}, exhibiting great gaps between real-world paper figures.

\subsubsection{Analysis}
We investigate how different subject domains affect mathematical reasoning ability using pairs of questions and answers (QA). We focus on six domains with more than 5K samples each. From each domain, we randomly choose a subset of 5K samples to ensure fairness in comparison. We then fine-tune the Qwen-VL-Chat base model using QA pairs from each domain and observe how it affects the model's accuracy compared to its original state.
Figure~\ref{fig:domain_analysis} demonstrates the relative accuracy changes (i.e., $\frac{\text{Accuracy after Fine-tuning}}{\text{Original Accuracy}} - 1 $) after training the model on QA pairs from each domain. Our findings reveal several key points: (i) QA pairs from the Computer Science (CS) domain are highly effective for improving mathematical reasoning ability, achieving a notable 27.09\% relative improvement. We attribute this to the compositional nature of the CS area. (ii) The most beneficial domain varies depending on the specific task. For instance, QA pairs from astrophysics domains enhance geometry problem-solving, while those from Condensed Matter improve performance in math word problems. (iii) Most domains hurt the Figure QA task. This suggests that synthetic Figure QA might not be the best benchmark for assessing realistic reasoning ability.
These findings underscore the efficacy of generated QA pairs and offer valuable insights for future research, such as adjusting task-specific weights in the dataset accordingly.

%% file: 03_method.tex
\subsection{Benchmarking LVLMs on ArXivCap}
\label{subsec:exp_arxivcap}
\subsubsection{Evaluated Tasks}
\label{subsubsec:evaluated_task}
Four vision-to-text tasks to benchmark LVLMs' ability to comprehend scientific figures.
\paragraph{Single-Figure Captioning} 
Similar to the traditional image captioning setup~\citep{lin2014mscoco}, single-figure captioning requires the model to generate a caption for the given single figure. 
The captions generated by the model are expected to encapsulate the nuanced details within these figures, including numbers and mathematical formulas, presenting a unique challenge for models to identify and articulate these elements accurately.
Formally, given an image-caption pair $(I, C)$, the LVLM $\mathcal{M}$ is asked to generate the caption given an instruction prompt $P_s$ to hint the goal of scientific captioning:
\begin{equation*}
    \hat{C} = \mathcal{M} (I, P_s),
\end{equation*}
where $\hat{C}$ would be evaluated according to the ground-truth $C$.

\paragraph{Multiple-Figure Captioning}
We introduce a more intricate challenge involving applying reasoning across multiple images. This task, termed Multiple-Figure Captioning, necessitates the model to craft a comprehensive summary caption for subfigures. 
As exemplified in Figure~\ref{fig:chunk_example}, the model is tasked with generating an overarching caption for two or more subfigures, leveraging visual clues to draw comparisons and formulate summary captions.
Formally, given a list of figures $L = \left( I_1, \ldots, I_n\right)$, the model is asked to generate the ground-truth main caption $C$ by considering all the semantics in the figures with a task prompt $P_m$:
\begin{equation*}
    \hat{C} = \mathcal{M} ( L, P_m ) = \mathcal{M} ( I_1, \ldots, I_n, P_m).
\end{equation*}

\paragraph{Contextualized Captioning}
Inspired by the evolving in-context learning capabilities of LLMs~\citep{brown2020language,icl_survey}, we introduce a contextualized captioning task to examine the in-context learning ability of LVLMs. 
In this task, the model is presented with a set of figure-caption pairs, and its goal is to generate a caption for a given image based on the provided demonstrations.
Given a sequential image-captions pairs $S = \{ ( I_i, C_i)  \}_{i=1}^n$ consisting of $n$ pairs of image $I_i$ and the corresponding $C_i$, the contextualized image captioning task can be formalized as follows:
\begin{equation*}
    \hat{C_n} = \mathcal{M} ( I_1, C_1, \ldots, I_{n-1}, C_{n-1}, I_n, P_c).
\end{equation*}
The model is supposed to leverage the context history to enhance the accuracy and coherence of the generated caption.

\paragraph{Title Generation}
This task requires a nuanced understanding of figures and captions to distill essential observations into a high-level summary of the presented results for LVLMs.
Specifically, instead of producing the captions for the figures, this task requires the model to connect different figures and corresponding captions to infer the paper title.
Let $S = \{ (I_i, C_i) \}_{i=1}^m$ be a sequence of $m$ figure-caption pairs in the extracted paper. 
Note that $I_i$ could be a single figure or a multiple-figure, and we reuse $I_i$ for simplicity here.
The title generation asks $\mathcal{M}$ to generate the title for the paper given a task prompt $P_t$:
\begin{equation*}
   \hat{T} = \mathcal{M} ( I_1, C_1, \ldots, I_{m}, C_m, P_t ) .
\end{equation*}
The prediction $\hat{T}$ is evaluated by comparing it to the original title $T$.

%% file: tables/single_cap.tex
\begin{table}[t!]
    \centering
    \resizebox{\linewidth}{!}{
    \begin{tabular}{@{}l|ccc@{}}
    \toprule
    \textbf{Model}     &  BLEU-2 & ROUGE-L & BERT-S\\
    \midrule 
     BLIP-2-OPT-6.7B    & 2.1 & 7.1 & 81.1 \\ 
     InstructBLIP-Vicuna7B  & 3.7& 10.1 & 83.3 \\ 
     LLaVA-v1.5-7B    & 2.3 & 10.6 & 83.0 \\ 
     LLaVA-v1.5-13B    &2.6  & 10.7&  83.3 \\ 
      \midrule 
     OpenFlamingo-9B    & 5.7 &  9.9 & 82.4  \\ 
     IDEFICS-Instruct-9B & 2.5 & 9.1 & 83.5 \\ 
     Qwen-VL-Chat-7B &4.4 &11.1 & 81.8\\ 
    Qwen-VL-Chat-7B$_\text{ArXivCap}$ & \textbf{8.9} & \textbf{15.8} & \textbf{83.3} \\ 
     \midrule
     \textcolor{gray}{Gemini 1.0 Pro Vision} & \textcolor{gray}{5.6} & \textcolor{gray}{14.5} & \textcolor{gray}{82.2}\\ 
     
      \textcolor{gray}{GPT-4V} & \textcolor{gray}{5.5} &\textcolor{gray} {14.2}  & \textcolor{gray}{83.3}\\  
     \bottomrule
    \end{tabular}}
    \caption{Evaluation results of single figure captioning. \textcolor{gray}{Grey} results are obtained from a 500-sample subset. Despite most LVLMs struggle to produce high-quality captions of scientific figures, training with ArXivCap significantly boosts the performance.}
    \label{tab:single_cap_ret}
\end{table}

%% file: tables/cap_with_title_abstract.tex
\begin{table}[t!]
    \centering
    \resizebox{\linewidth}{!}{
    \begin{tabular}{@{}l|ccc@{}}
    \toprule
    \textbf{Model}     &  BLEU-2 & ROUGE-L & BERT-S\\
    \midrule 
     Qwen-VL-Chat-7B &4.4 &11.1 & 81.8\\ 
      \quad + Title &5.7	&13.1&	81.6 \\ 
      \quad + Title and Abstract 	 & 6.0	 & 12.7	 & 81.4 \\ 
     \midrule 
    Qwen-VL-Chat-7B$_\text{ArXivCap}$ & 8.9 & 15.8 & 83.3 \\ 
     \quad + Title & \textbf{12.9}&	\textbf{18.6}	& \textbf{83.8} \\ 
      \quad + Title and Abstract &	12.7&	18.5&	83.8   \\      
     \bottomrule
    \end{tabular}}
    \caption{Evaluation results of single figure captioning with paper meta information.}
    \label{tab:cap_with_meta_ret}
\end{table}

%% file: tables/main_ret.tex
\begin{table*}[tbh!]
\small 
\centering
\resizebox{\linewidth}{!}{
\begin{tabular}{@{}l|ccc|ccc|ccc@{}}
\toprule
\multirow{2}{*}{\textbf{Model}} & 
\multicolumn{3}{c}{\textbf{Multiple-Figure Captioning}}  & \multicolumn{3}{c}{\textbf{Contextualized Captioning}} & \multicolumn{3}{c}{\textbf{Title Generation}} \\
                       & BLEU-2 & ROUGE-L & BERT-S & BLEU-2 & ROUGE-L & BERT-S & BLEU-2 & ROUGE-L & BERT-S \\

\midrule 
OpenFlamingo-9B  &     3.7  &  11.3     &  81.9           &    20.0  &     20.5  &         83.7  &    2.7   &   17.7    &  82.7      \\
IDEFICS-Instruct-9B &   3.6   & 10.8      &   82.8         & \textbf{20.7}   &     \textbf{22.6}  &     \textbf{85.7}        &    3.5    &     18.4 &     85.8    \\

Qwen-VL-Chat-7B   & 3.0     & 7.2   &  79.7          & 17.0   &   22.1  &  85.0         &  2.6    &     15.8  &   85.1         \\
 Qwen-VL-Chat-7B$_\text{ArXivCap}$ &  \textbf{10.6}     & \textbf{18.0}      & \textbf{83.6}           &    16.1    &    21.2     &        84.8     &     \textbf{6.7} &   \textbf{23.5}    & \textbf{86.8}      \\
\midrule
\textcolor{gray}{Gemini 1.0 Pro Vision}   &  \textcolor{gray}{6.1}    &   \textcolor{gray}{16.2}   &   \textcolor{gray}{83.1}     &  \textcolor{gray}{10.2}     &  \textcolor{gray}{20.2}       & \textcolor{gray}{84.5}           &   \textcolor{gray}{5.7}   &   \textcolor{gray}{21.8}  & \textcolor{gray}{85.9}     \\
\textcolor{gray}{GPT-4V}  &  \textcolor{gray}{5.7}   &  \textcolor{gray}{14.7}    &   \textcolor{gray}{83.0}         &    \textcolor{gray}{9.6}  &\textcolor{gray}{20.1}       & \textcolor{gray}{84.7}          &   \textcolor{gray}{4.0}   &   \textcolor{gray}{20.2}  &   \textcolor{gray}{86.0}     \\
\bottomrule
\end{tabular}
}
\caption{Evaluation results of three newly defined tasks. The best results are highlighted in \textbf{bold}.}
\label{tab:ret_three_tasks}
\end{table*}

%% file: 10_conclusion.tex
\section{Conclusion}
\label{sec:conclusion}
Our work introduces Multimodal ArXiv, comprising ArXivCap and ArXivQA, aims at advancing the scientific comprehension of LVLMs. 
Experiments show that fine-tuning on ArXivQA notably enhances LVLMs' mathematical reasoning capabilities. 
Moreover, our comprehensive evaluations across four vision-to-text tasks on ArXivCap underscore the challenges in understanding scientific figures for LVLMs, while highlighting the substantial improvements achieved by in-domain training. Our error analysis offers valuable insights for the ongoing development of LVLMs.



\section*{Limitations}
Our study has several limitations worth noting. Firstly, our exploration may not encompass the full spectrum of LVLMs due to the rapid evolution of architectures and training methodologies such as parameter-efficient tuning~\citep{hu2021lora,ma2024paramtuning}.
Nevertheless, we believe our dataset could still be effective for other LVLMs and the findings are generalizable.
We show that our ArXivQA dataset could also boost LLaVA-series models across scientific understanding benchmarks in Appendix~\ref{apx:llava}.
Secondly, our Multimodal ArXiv dataset sources from ArXiv papers due to their accessibility and open-source licenses. This approach may overlook the diversity of disciplines and data modalities present in the broader scientific literature. 
Future research could incorporate a broader range of datasets and domains to enrich the coverage of scientific knowledge, and explore dynamic data selection methods to improve performance and sample efficiency~\citep{li-etal-2021-dynamic,chen2024vlm_select}.


%% file: ack.tex
\section*{Acknowledgements}
We would like to thank all the anonymous reviewers for their insightful comments and suggestions, which helped us improve the quality and clarity of this work. 
We are particularly grateful to Shuhuai Ren for his valuable feedback in preparing the manuscript.
This research
was supported in part by the joint research scheme of the
National Natural Science Foundation of China (NSFC) and
the Research Grants Council (RGC) under grant number
N HKU714/21.

%% file: 12_appendix.tex
\appendix

\section{Details of Multimodal ArXiv}
\label{apx:arxiv_cap}

\subsection{Caption Cleaning}
\label{apx:caption_clean}

We apply a Python tool to clean the original caption and Table~\ref{tab:pylatexenc_clean} illustrates the caption before and after cleaning. 
\input{tables/pylatexenc_clean}

\subsection{Illustration Cases of Multimodal ArXiv}
\label{apx:case_illustrations}
\begin{figure}[htb!]
    \centering
    \includegraphics[width=\linewidth]{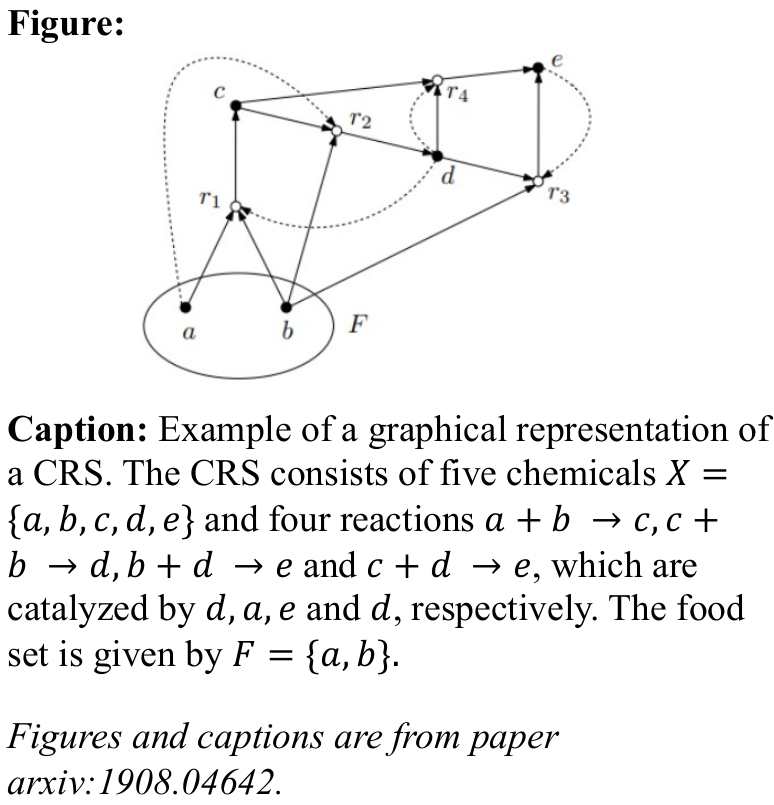}
    \caption{A single-figure caption pair in our ArXivCap dataset.}
\label{fig:single_figure_pair}
\end{figure}

\begin{figure*}[t!]
    \centering
    \includegraphics[width=0.85\linewidth]{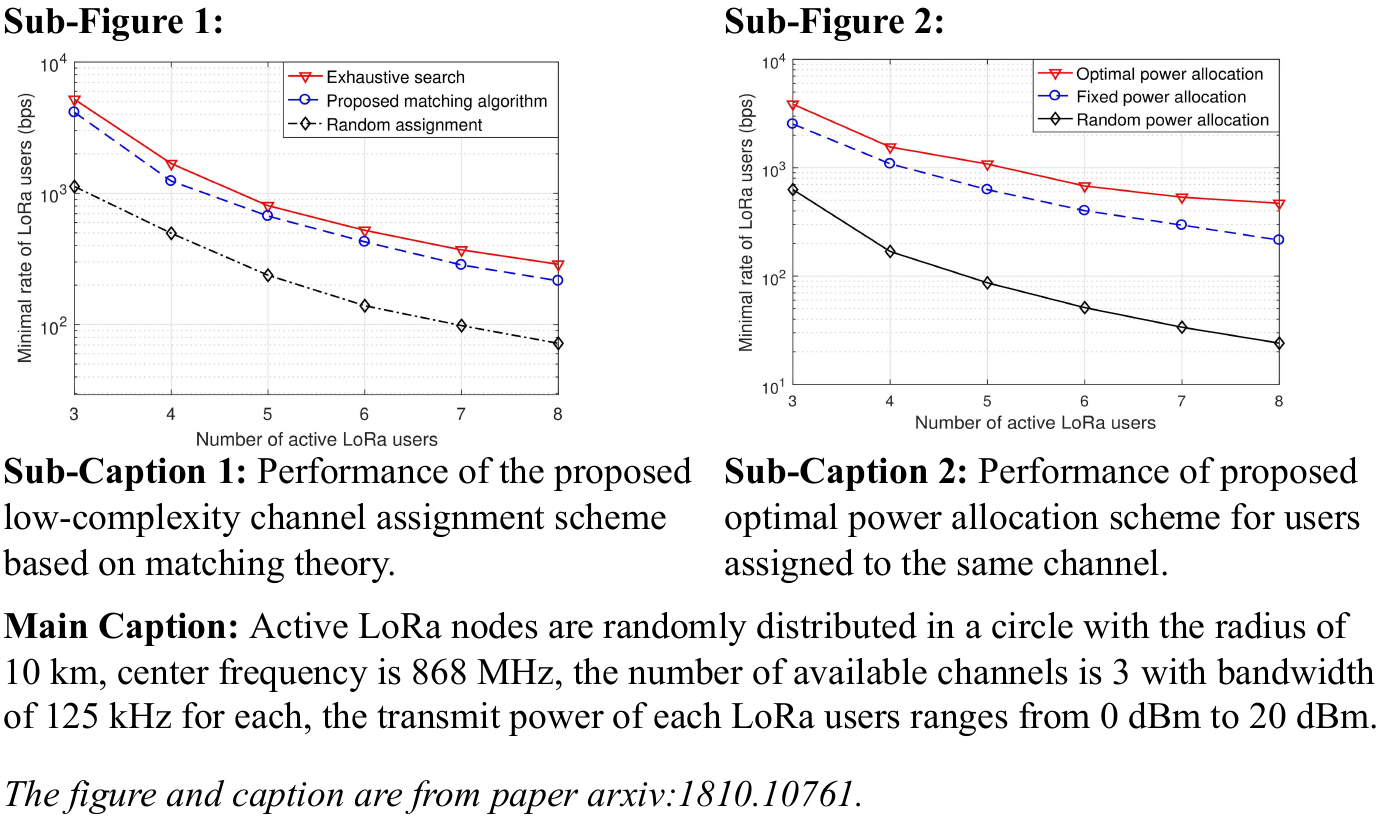}
    \caption{A multiple-figure caption pair in our ArXivCap dataset.}
    \label{fig:multi_figure_case}
\end{figure*}

\begin{figure*}[tbh!]
    \centering
    \includegraphics[width=0.85\linewidth]{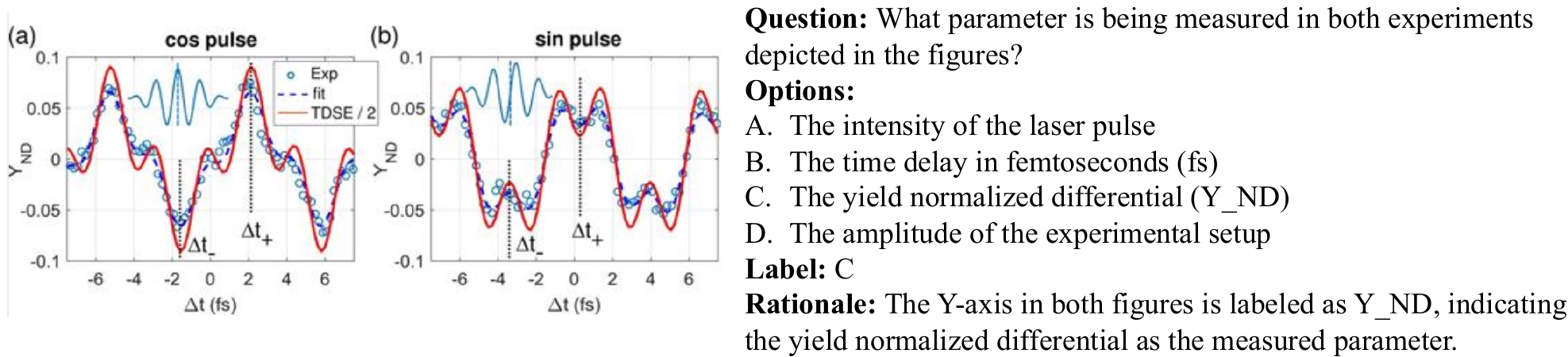}
    \caption{A case from our ArXivQA dataset.}
    \label{fig:qa_case_1}
\end{figure*}

\begin{figure*}[tbh!]
    \centering
    \includegraphics[width=0.85\linewidth]{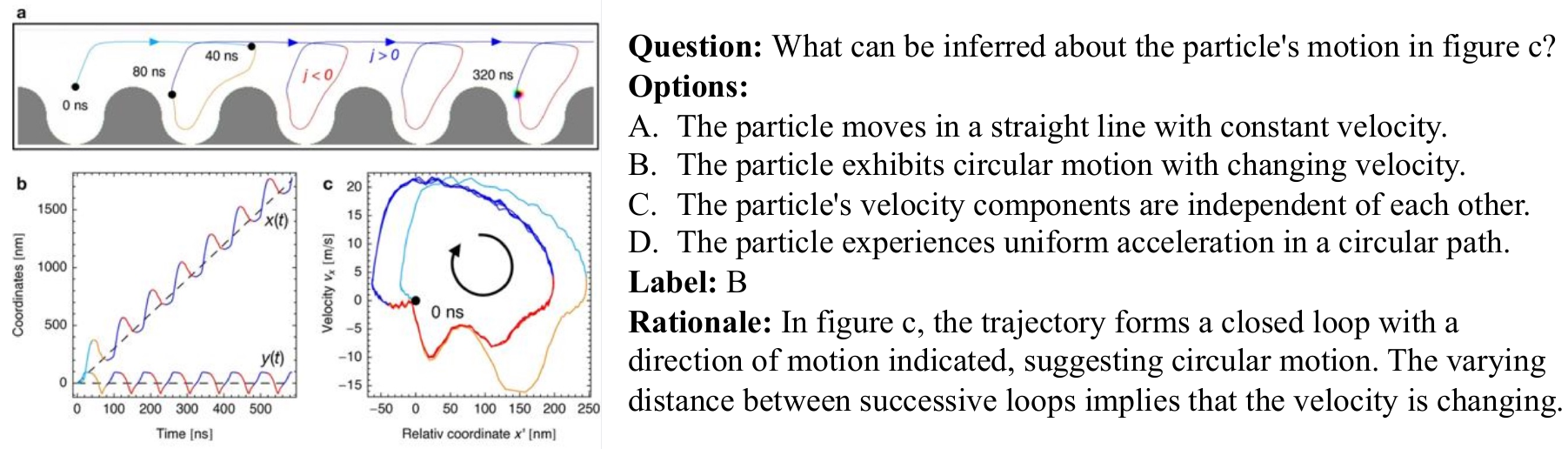}
    \caption{A case from our ArXivQA dataset.}
    \label{fig:qa_case_2}
\end{figure*}

\begin{figure*}[tbh!]
    \centering
    \includegraphics[width=0.85\linewidth]{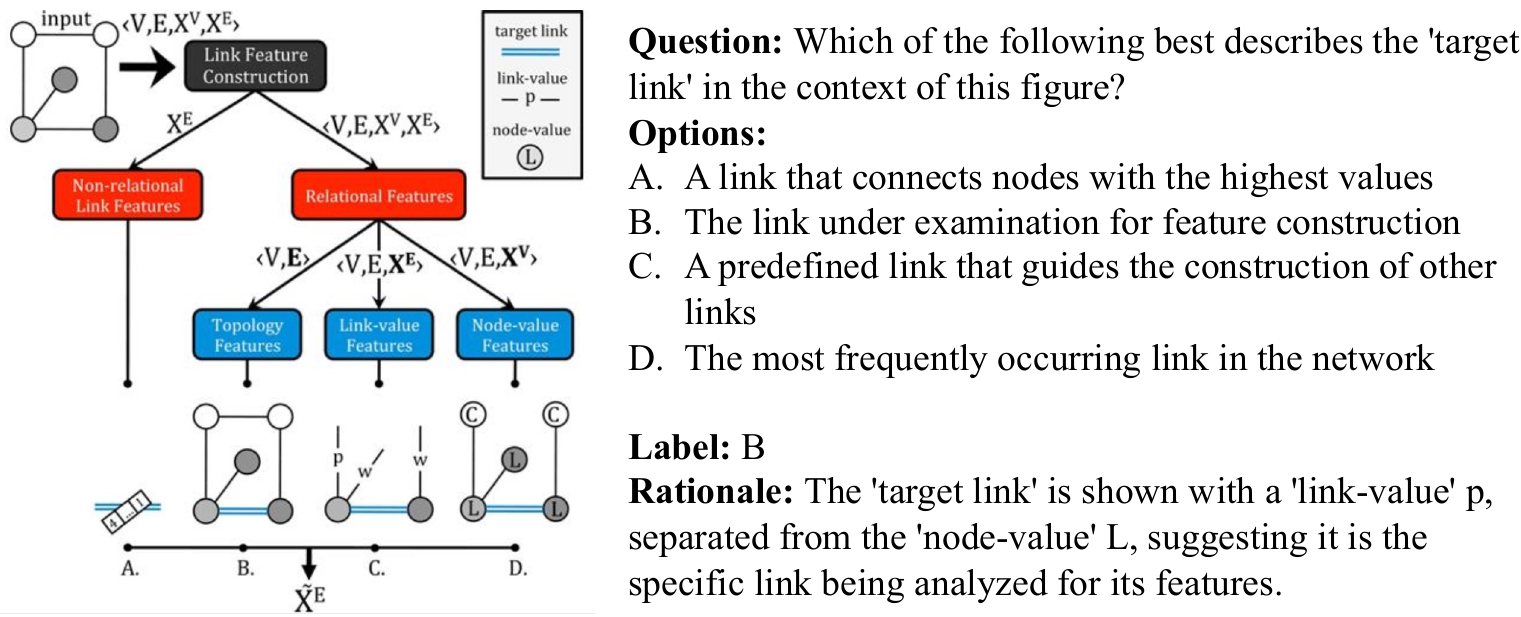}
    \caption{A case from our ArXivQA dataset.}
    \label{fig:qa_case_3}
\end{figure*}

\begin{figure*}[tbh!]
    \centering
    \includegraphics[width=0.85\linewidth]{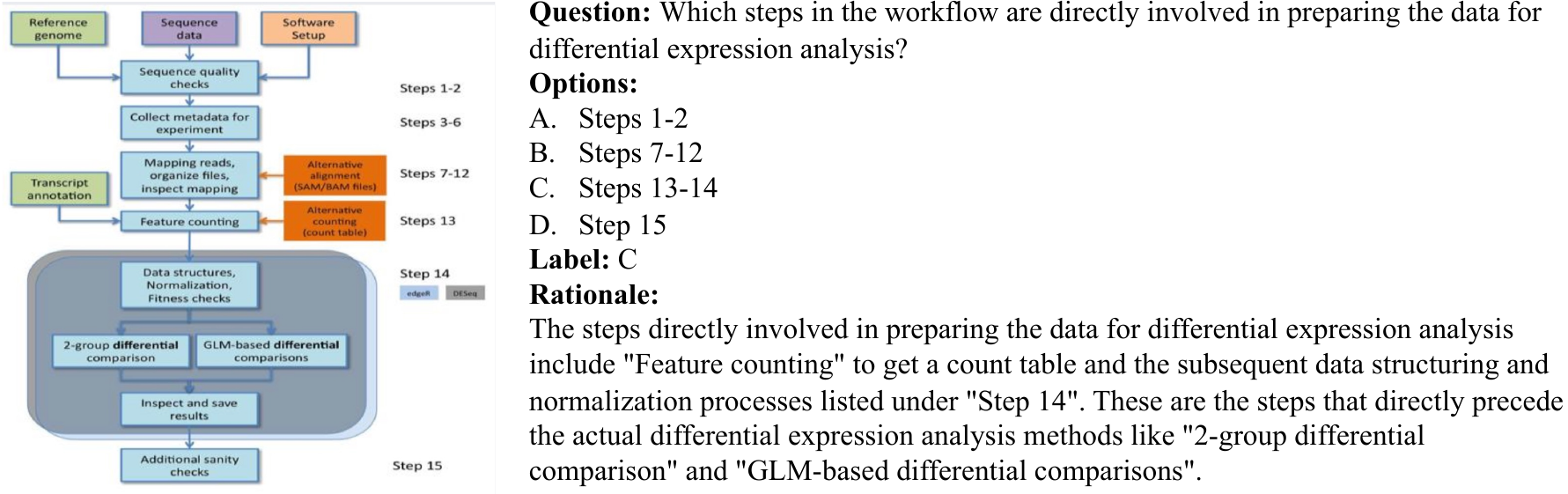}
    \caption{A case from our ArXivQA dataset.}
    \label{fig:qa_case_4}
\end{figure*}

We provide illustrated cases from our dataset for a better understanding.
Figure~\ref{fig:single_figure_pair} demonstrates a typical single figure-caption pair, and Figure~\ref{fig:multi_figure_case} shows the multiple figure-caption case.
Figure~\ref{fig:qa_case_1}, \ref{fig:qa_case_2}, \ref{fig:qa_case_3} and \ref{fig:qa_case_4} illustrate the cases from our ArXivQA dataset, covering different figure types and containing diverse questions.

\subsection{Caption Word Cloud}
\label{apx:caption_word_cloud}

\begin{figure}[t!]
    \centering
    \includegraphics[width=\linewidth]{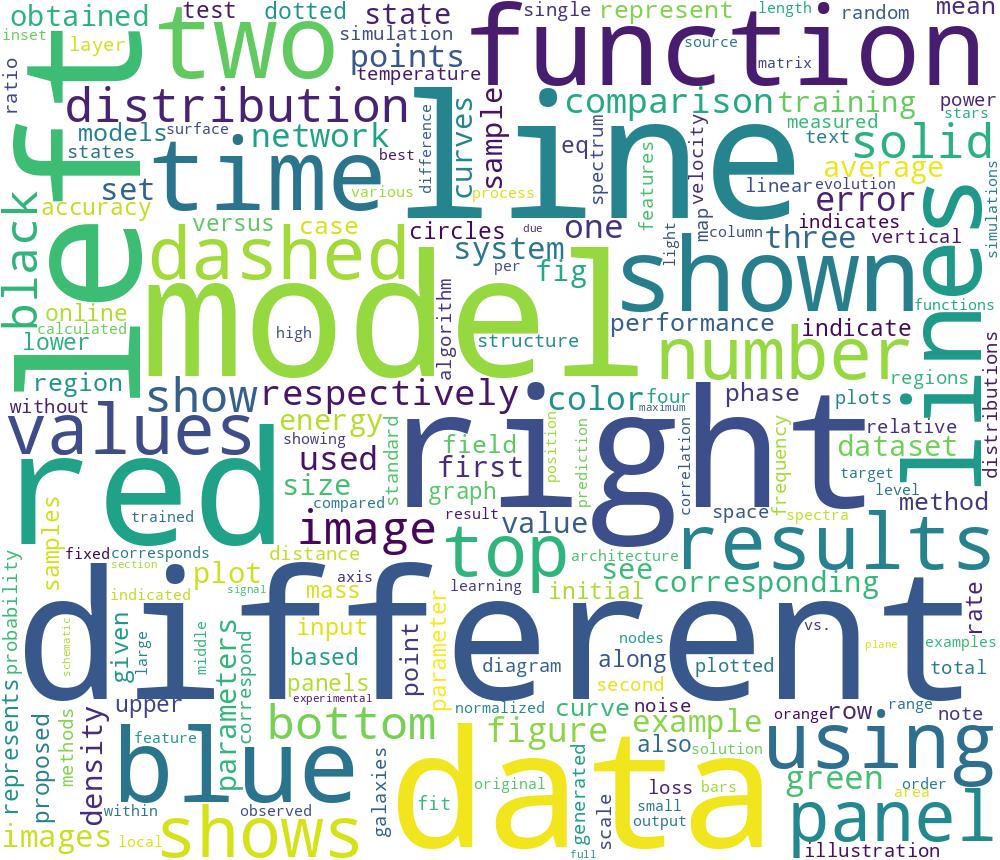}
    \caption{Word cloud visualization of captions in our ArXivCap dataset.}
    \label{fig:word_cloud}
\end{figure}

We visualize the word cloud of captions in our ArXivCap dataset in Figure~\ref{fig:word_cloud}. It can be seen that the captions have a diverse vocabulary for describing the different figures in the academic papers.

\input{tables/domain-full-name}

\input{tables/arxivqa_template}
\subsection{ArXivQA Prompting Template}
\label{apx:prompt_template}
The prompt used to query GPT-4V is provided in Table~\ref{tab:prompt_for_ArXivqa}.

\subsection{Quality Analysis of ArXivQA}
\input{tables/arxivqa_quality_analysis}
To evaluate the quality of ArXivQA, we manually assess it from six different aspects. We develop a quality examination guideline for annotators, as shown in Table~\ref{tab:quality_analysis_ArXivqa}, which addresses various aspects of the QA pairs.
We sample 100 examples and ask four authors to conduct the quality analysis. The four authors are divided into two groups, with each group tasks with evaluating 50 examples across six sub-aspects, according to the grading protocol. 

The evaluation results are presented in Table~\ref{tab:quality_analysis_ArXivqa_result}.
We find that most samples feature clear, high-quality images with clear and high-quality images, with unambiguous question and option descriptions. However, a small fraction of the generated questions may be unanswerable due to mis-recognizing elements in the figures, as reflected by lower factual alignment scores. Additionally, we consider samples with an aggregate score of 5 or higher from both annotators to be of sufficient quality. Under this stringent criterion, 79 out of 100 samples meet the threshold, demonstrating that the dataset's quality is generally satisfactory.

\section{Evaluation Details}
\label{apx:evaluation_details}
\subsection{Details of Evaluated Models}

\paragraph{BLIP2}~\citep{li2023blip2}, 
introduces a lightweight Q-Former designed to bridge modality gaps and leverages frozen LLMs. Leveraging LLMs, BLIP-2 can conduct zero-shot image-to-text generation using natural language prompts. We select the \texttt{BLIP2-OPT-6.7B} version for evaluation.\footnote{\url{https://huggingface.co/Salesforce/blip2-opt-6.7b}}

\paragraph{InstructBLIP}~\citep{dai2023instructblip} employs an instruction-aware visual feature extraction module based on BLIP2~\citep{li2023blip2} and is trained with unified multimodal instruction tuning datasets. We choose \texttt{InstructBLIP-Vicuna-7B} for evaluation.\footnote{\url{https://huggingface.co/Salesforce/instructblip-vicuna-7b}}.

\paragraph{LLaVA}~\citep{liu2023llava},
adopts Vicuna models as the backbone LLM and is trained on the ChatGPT/GPT-4 generated instruction tuning dataset.
LLaVA-v1.5~\citep{liu2023llava15} improves on LLaVA models by employing curated task datasets and an enhanced modality alignment module.
We evaluate both \texttt{LLaVA-v1.5-7B}\footnote{\url{https://huggingface.co/liuhaotian/llava-v1.5-7b}} and \texttt{LLaVA-v1.5-13B}.\footnote{\url{https://huggingface.co/liuhaotian/llava-v1.5-13b}}

\paragraph{Flamingo}~\citep{Alayrac2022FlamingoAV} pioneers the development of LVLMs by introducing a cross-gated layer for LLMs to produce visual-grounded text. The training dataset
consists of interleaved visual data and text from the web pages, enabling it to generate free-form text as the output. We select the open-source implementation \texttt{OpenFlamingo-9B}~\citep{awadalla2023openflamingo} for evaluation.\footnote{\url{https://huggingface.co/openflamingo/OpenFlamingo-9B-vitl-mpt7b}}

\input{tables/GPT-4_Score}

\paragraph{IDEFICS} is another open-sourced implementation of Flamingo~\citep{Alayrac2022FlamingoAV}. Trained on publicly available image-text alignment pairs and instruction tuning datasets, it demonstrates comparable results with the original closed-source model on various image-text benchmarks. We select the \texttt{IDEFICS-Instruct-9B} for evaluation.\footnote{\url{https://huggingface.co/HuggingFaceM4/idefics-9b-instruct}}.

\paragraph{Qwen-VL-Chat}~\citep{Qwen-VL} is
a bilingual LVLM that supports both English and Chinese built on the Qwen LLM~\citep{qwen}. 
During the training phase, Qwen-VL-Chat adopts a packing strategy to create multiple images as inputs,
improving its ability to understand the vision context. We select the \texttt{Qwen-VL-Chat-7B} for evaluation.\footnote{\url{https://github.com/QwenLM/Qwen-VL}}

\paragraph{GPT-4V}
~\citep{gpt4v}, the proprietary vision-language models developed by OpenAI, which are shown to be powerful on various multi-modal tasks~\citep{yang2023dawn}. The API version we queried is \texttt{gpt-4-vision-preview}.

\paragraph{Bard}~\citep{bard}, a commercial LVLM developed by Google. We utilize the unofficial API\footnote{\url{https://github.com/dsdanielpark/Bard-API}} querying the model with our task prompts, accessed on \texttt{2023-11-17}.

\paragraph{Gemini 1.0 Pro Vision}~\citep{reid2024gemini}, a upgraded LVLM by Google. We utilize the official API querying the model with our task prompts, accessed on \texttt{2024-05-20}.

\subsection{Task Prompts}
We evaluate all the models with the same task prompts in our experiments, and the prompts for our four tasks are listed below:

\noindent\textbf{Single-Figure Captioning}:  \texttt{Create a caption for the provided figure.}

\noindent\textbf{Multiple-Figure Captioning} \texttt{Create a caption for the provided figures.}

\noindent\textbf{Contextualized Captioning}: We reuse the prompts in previous captioning tasks depending on the current figure type.

\noindent\textbf{Title Generation}: \texttt{
According to the figures and captions, generate a title for this paper. Title:}

\subsection{GPT-4 Evaluation of Caption}
\label{apx:gpt4_eval}

In addition to BLEU-2, ROUGE-L, and BERT-S, we also utilize GPT-4 to evaluate a sample of 500 generated captions. Specifically, we employ GPT-4 for the evaluation of single-figure caption tasks following FairEval~\citep{wang2023large}.
The template for prompting GPT-4 to evaluate generated captions is presented in Table~\ref{tab:prompt_for_gpt4_score_caption}. GPT-4 is asked to perform an analysis and then produces a quality score, which is subsequently mapped to a scale from 1 to 5.
The results are presented in Table~\ref{tab:gpt4_score_single_caption}. We observe that the ROUGE-L metric exhibits the highest correlation with the GPT-4 Score (Pearson r = 0.91), followed by BLEU-2 (Pearson r = 0.64). BERT-S instead demonstrates a moderate correlation (Pearson r = 0.39).
The uniformly low GPT-4 scores across all models suggest that they struggle to produce satisfactory captions, which is consistent with the findings in our main paper. Notably, training on ArXivCap results in a significant 12\% improvement in the GPT-4 score compared to the original Qwen-VL-Chat model, leading to the most favorable outcomes in this evaluation.

\section{Error Analysis}
\label{apx:case}
\begin{figure*}[t!]
    \centering
    \includegraphics[width=0.95\linewidth]{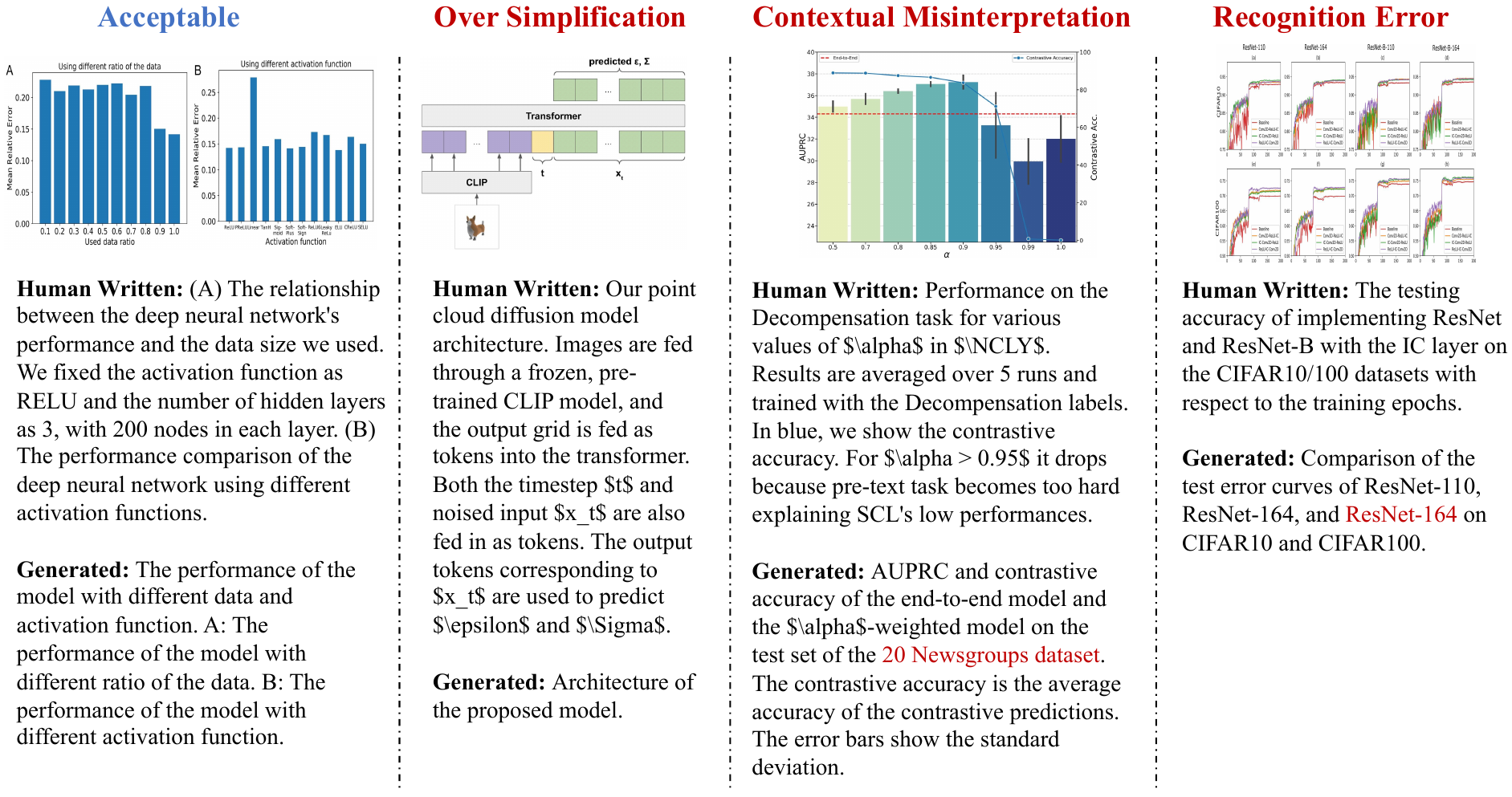}
    \caption{Illustration of acceptable and three error types of generated captions.}
    \label{fig:caption_type}
\end{figure*}

\begin{figure}[t!]
    \centering
    \includegraphics[width=0.85\linewidth]{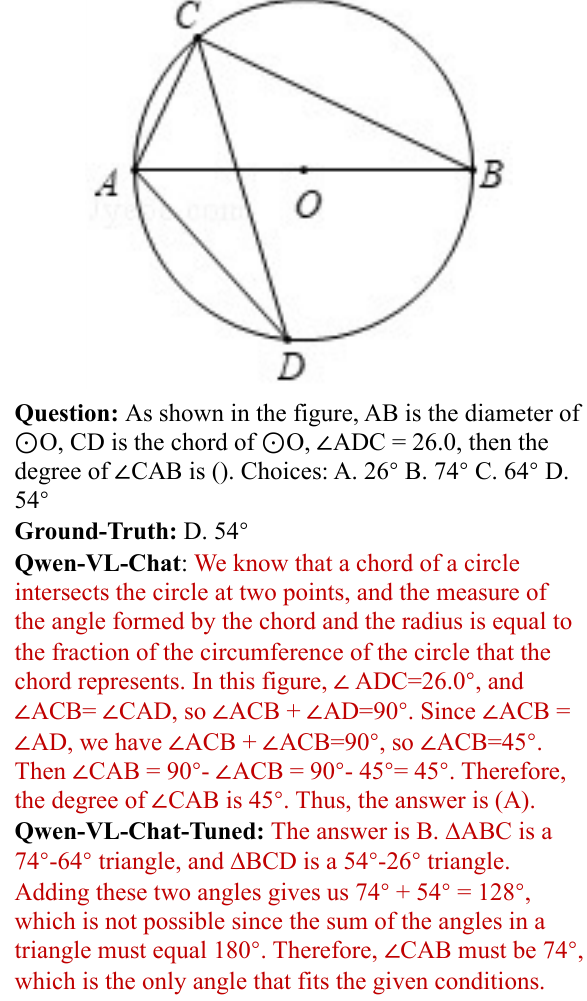}
    \caption{A failure case on the geometry problem-solving task.}
\label{fig:geometry_fail}
\end{figure}
\subsection{Caption Type Illustration} 
\label{apx:caption_type}
We illustrate captions of four types in our main paper in Figure~\ref{fig:caption_type}. The \emph{Acceptable} caption provides a comprehensive description of the figure presented. The \emph{oversimplified} caption is too short compared with the original human-written caption. 
Furthermore, as shown in the third block in Figure~\ref{fig:caption_type}, \emph{Contextual Misinterpretation} refers to captions with unmentioned content in the figure, such as the dataset colored in red.
\emph{Recognition Error} denotes the model wrongly identified the number or text in the figure, such as the misidentified model name in the last block of Figure~\ref{fig:caption_type}.

\subsection{Failure Sample of MathVista}
\label{apx:failure_mathvista}
Figure~\ref{fig:geometry_fail} shows a challenging geometry mathematic reasoning problem where both models fail to produce the correct answer.
Echoing the quantitative results in our main paper, we believe future studies can incorporate more focused corpus for enhancing the geometry and mathematical reasoning ability of LVLMs.

\section{Results with LLaVA Backbone}
\label{apx:llava}
\input{tables/llava_ret}

We investigate whether ArXivQA could also enhance other LVLMs, such as LLaVA models~\citep{liu2023llava}. To maintain model performance, we mix our ArXivQA dataset with the LLaVA SFT 665K-instruction tuning dataset. The LLaVA-v1.5-7B is adopted as the backbone and the model is trained following the original recipe. The results on various benchmarks are listed in Table \ref{tab:llava_performance}.
We find that not only the scientific reasoning performance is improved on multimodal reasoning tasks (MathVista \citep{mathvista}, MMMU \citep{mmmu}, and ScienceQA \citep{lu2022scienceqa}), but the overall capability on MM-Vet \citep{yu2024mmvet} is also boosted. Together with our results using Qwen-VL-Chat, these findings indicate that our ArXivQA dataset can enhance different model backbones and is beneficial across various benchmarks.

%% file: tables/pylatexenc_clean.tex
\begin{table}[tbh!]
    \centering
    \begin{tabular}{p{3cm}  p{3cm}}
    \toprule
       Before Cleaning & After Cleaning \\
       \midrule
       A 1995 Hale Telescope H$\alpha$ image of the Guitar Nebula (20 angstrom filter at 6564 angstroms). The cometary neck connecting to a spherical bubble are clearly evident. Credit: \textbackslash cite\{cha02\}. &
        A 1995 Hale Telescope H$\alpha$ image of the Guitar Nebula (20 angstrom filter at 6564 angstroms). The cometary neck connecting to a spherical bubble are clearly evident. Credit: \textless cit.\textgreater. \\ 
       \midrule
       As Fig. \textbackslash ref\{z0\} except at $z\sim 6$ ($z=4.37$ in the EdS model). &
    As Fig. \textless ref\textgreater~except at $z\sim 6$ ($z=4.37$ in the EdS model). \\ 
 
       \bottomrule
    \end{tabular}
    \caption{Caption before and after cleaning using pylatexenc.}
    \label{tab:pylatexenc_clean}
\end{table}

%% file: tables/domain-full-name.tex
\begin{table}[tbh!]
    \centering
    \small 
    \resizebox{\linewidth}{!}{
    \begin{tabular}{l|c}
    \toprule
       Domain  & Full Name \\
       \midrule     
        dg-ga & Differential Geometry \\
        acc-phys & Accelerator Physics  \\
        solv-int & Exactly Solvable and Integrable Systems  \\
        q-alg &  Quantum Algebra and Topology \\
        atom-ph &  Atomic, Molecular and Optical Physics \\
        alg-geom & Algebraic Geometry  \\
        comp-gas &  Cellular Automata and Lattice Gases \\
        supr-con & Superconductivity  \\
        chem-ph &  Chemical Physics \\
        mtrl-th & Materials Theory  \\
        adap-org &  Adaptation, Noise, and Self-Organizing Systems \\
        patt-sol & Pattern Formation and Solitons  \\
        chao-dyn &  Chaotic Dynamics \\
        cmp-lg &  Computation and Language \\
        econ & Economics  \\
        hep-lat & High Energy Physics - Lattice  \\
        nucl-ex & Nuclear Experiment  \\
        q-fin &  Quantitative Finance \\
        math-ph &  Mathematical Physics \\
        nucl-th &  Nuclear Theory \\
        gr-qc & General Relativity and Quantum Cosmology  \\
        hep-ex & High Energy Physics - Experiment  \\
        hep-th & High Energy Physics - Theory \\
        nlin & Nonlinear Sciences \\
        hep-ph & High Energy Physics - Phenomenology \\
        q-bio & Quantitative Biology \\
        quant-ph & Quantum Physics \\
        eess & Electrical Engineering and Systems Science \\
        stat & Statistics \\
        astro-ph & Astrophysics \\
        physics & Physics  \\
        cond-mat & Condensed Matter \\
        math & Mathematics \\
        cs & Computer Science \\
       \bottomrule
    \end{tabular}}
    \caption{Name of each domain.}
    \label{tab:domain-full-name}
\end{table}

%% file: tables/arxivqa_template.tex
\begin{table}[t!]
    \centering
    \tiny 
    \begin{tcolorbox}
Multiple-choice Question Answer Pairs Generation for Scientific Figures

\textbf{Guideline}

The goal of this task is to create answerable multiple-choice questions based on figures from scientific papers, to improve the ability of a large vision language model.

The questions should be challenging, and require college-level reasoning. The type of questions should be diverse. The question should be answerable based on the figure. The answer should be one of the answer choices. The answer choices should be plausible and challenging.

\textbf{Format}

Below is an example of the format of the input and output for the task.

\textbf{Input}

Figures: [Figures input in the task]

\textbf{Output}

Question: [Question]

Answer Options: [Answer choices, a bullet list.]

Correct Choice: [Correct answer choice, e.g., A]

Rationale: [Rationale for the correct answer, explain why the answer is correct]
    \end{tcolorbox}
    \caption{Prompt used for GPT-4V to generate QA pairs based on scientific figures.}
    \label{tab:prompt_for_ArXivqa}
\end{table}

%% file: tables/arxivqa_quality_analysis.tex
\begin{table*}[t!]
    \centering
    \tiny  
    \begin{tcolorbox}
1. Ensuring Factual Integrity and Clear Presentation:\\
- Factual Alignment: Ensure questions and options are grounded in accurate reflections of the chart data.\\
- Visual Clarity: Maintain high-resolution charts to ensure that all pertinent details are discernible.\\
- Unambiguous Textual Information: Employ precise and unambiguous language to formulate questions and answers, thereby mitigating potential misinterpretations.\\
2. Ensuring Relevance and Integrated Comprehensiveness:\\
- Question and Option Relevance: Charts must align with their questions, and all options should be applicable and relevant to the given data.\\
- Comprehensive Integration: Guarantee the provision of comprehensive information necessary for the interpretation of the chart and the resolution of the question, ensuring a cohesive amalgamation of textual and visual data.\\ 
3. Promoting Fairness and Avoiding Bias:\\
- Equitable Content: Strive for impartiality in the dataset to prevent bias and ensure fair representation of diverse groups and perspectives.\\

Grading Protocol:\\
Each criterion is to be rigorously evaluated for each dataset entry. The assessment is to be conducted on a qualitative scale with three distinct levels: High, Medium, and Low. These levels will denote the degree of conformity to the respective criterion:\\
- 1: High: The dataset entry exhibits exemplary adherence to the evaluation criterion, demonstrating a robust and comprehensive alignment with the specified standard.\\
- 0.5: Medium: The dataset entry meets the evaluation criterion to a moderate extent, indicating a satisfactory but not optimal congruence with the standard.\\
- 0 Low: The dataset entry falls short of the evaluation criterion, signaling a need for significant improvements to meet the standard.
    \end{tcolorbox}
    \caption{Annotator guideline of ArXivQA manual quality examination.}
    \label{tab:quality_analysis_ArXivqa}
\end{table*}

\begin{table}[t!]
    \centering
    \small 
    \begin{tabular}{@{}lc@{}}
         \toprule
         Aspect & Avg Score \\
         \midrule
        Factual Alignment & 0.6975 \\
        Visual Clarity & 0.9925 \\
        Unambiguous Textual Information & 0.9825 \\
        Question and Option Relevance & 0.9375 \\
        Comprehensive Integration & 0.905 \\
        Equitable Content & 1.0 \\
        \midrule
        Score Sum & 5.515 \\
        \bottomrule
    \end{tabular}
    \caption{Manual quality analysis of ArXivQA. The average scores for each aspect are presented.}
    \label{tab:quality_analysis_ArXivqa_result}
\end{table}

%% file: tables/GPT-4_Score.tex
\begin{table*}[t!]
    \centering
\tiny 
\begin{tcolorbox}
Annotation Instruction:\\
As an annotator, your role is to serve as an unbiased and objective judge in evaluating the accuracy of captions produced by a Large Vision-Language Model (LVLM) for scientific figures. These figures are extracted from academic papers, and to aid your assessment, we will provide you with the paper's title and abstract for necessary context.

You will be presented with the original caption—referred to as the 'ground truth'—and the LVLM generated caption, termed the 'prediction'. You could take into account the context given by the paper's title and abstract for background knowledge, comparing it critically with both captions.

In your assessment, please pay attention to the factual alignment, including but not limiting to the following aspects:\\
- Numerical data and statistics: Verify their accuracy and correspondence to the data presented in the figure.\\
- Symbols: Check for correct representation and usage in the context of the scientific subject matter.\\
- Factual content: Ensure all facts are consistent with those stated in the ground truth caption and the paper's content.\\

<title>{title}</title>\\
<abstract>{abstract}</abstract>\\
<ground truth>{gt}</ground truth>\\
<prediction>{pred}</prediction>\\

Compare the prediction to the ground truth, provide a brief analysis, and assign a score using one of the following quality labels: <Perfect>, <Good>, <Fair>, <Poor>, <Incorrect>.

Below we describe the detail criteria for score:
<Perfect>: The prediction is almost identical to the ground truth, with only minor, inconsequential differences that do not change the meaning. All numerical data, symbols, and factual content are accurate and consistent.\\
<Good>: The prediction is largely similar to the ground truth but has some noticeable differences that may slightly change the meaning. However, the core information is still correct, and the numerical data, symbols, and factual content are mostly accurate and consistent with the figure content.\\
<Fair>: The prediction captures the basic idea of the ground truth but has significant differences that change the meaning in a way that cannot be ignored. There may be some inaccuracies or inconsistencies in the numerical data, symbols, or factual content when compared to the figure content.\\
<Poor>: The prediction is related to the ground truth but has serious errors or omissions that significantly change the meaning. The numerical data, symbols, or factual content may be largely inaccurate or inconsistent with the figure content.\\
<Incorrect>: The prediction is completely different or irrelevant to the ground truth, with no similarities between the two. The numerical data, symbols, and factual content are entirely inaccurate or inconsistent with the figure content.

Give a brief analysis within 100 words and then output a quality label wrapped with "<>".
    \end{tcolorbox}
    \caption{Prompt template designed for GPT-4 to evaluate generated captions based on the paper title, abstract, and ground truth.}
    \label{tab:prompt_for_gpt4_score_caption}
\end{table*}

\begin{table}[t!]
    \centering
    \small  
    \resizebox{\linewidth}{!}{
    \begin{tabular}{@{}lcccc@{}}
         \toprule
         Model & BLEU-2 & ROUGE-L & BERT-S & GPT-4 Score \\
         \midrule
BLIP-2-OPT-6.7B & 1.5 & 6.6 & 81.3 & 1.18 \\
InstructBLIP-Vicuna7B & 3.5 & 10.3 & 83.6 & 1.48 \\
LLaVA-1.5-7B & 2.3 & 10.4 & 83.3 & 1.80 \\
LLaVA-1.5-13B & 2.7 & 11.0 & 83.6 & 1.69 \\
OpenFlamingo-9B & 5.8 & 10.3 & 82.7 & 1.52 \\
IDEFICS-Instruct-9B & 2.1 & 9.3 & 83.8 & 1.55 \\
Qwen-VL-Chat & 4.7 & 11.1 & 82.0 & 1.81 \\
Qwen-VL-Chat tuned w/ ArXivCap & 8.6 & 15.3 & 83.2 & 2.03 \\
         \bottomrule
    \end{tabular}}
    \caption{Results of 500 single-figure captions generated by various models.}
    \label{tab:gpt4_score_single_caption}
\end{table}

%% file: tables/llava_ret.tex
\begin{table*}[t!]
    \centering
    \small 
    \begin{tabular}{l|c c c c}
        \toprule
        Model & MathVista & MMMU(val) & ScienceQA(IMG only) & MM-Vet \\
        \midrule
        LLaVA-v1.5-7B & 26.6 & 35.3 & 66.8 & 30.5 \\
        \quad Original SFT +ArXivQA & \textbf{28.2} & \textbf{36.0} & \textbf{68.3} & \textbf{32.4} \\
        \bottomrule
    \end{tabular}
    \caption{After fine-tuning with a combination of ArXivQA and original SFT data, the LLaVA model shows boosted mathematical reasoning abilities across benchmarks.}
    \label{tab:llava_performance}
\end{table*}